%% file: main.tex
\documentclass[11pt,a4paper]{article}

\usepackage[utf8]{inputenc}
\usepackage[T1]{fontenc}
\usepackage{amsmath,amssymb,amsthm}
\usepackage{mathtools}
\usepackage{booktabs}
\usepackage{multirow}
\usepackage{graphicx}
\usepackage[hypertexnames=false]{hyperref}
\usepackage{cleveref}
\usepackage{enumitem}
\usepackage{xcolor}
\usepackage{natbib}
\usepackage[margin=1in]{geometry}

\newtheorem{theorem}{Theorem}

\newtheorem{definition}[theorem]{Definition}

\newcommand{\BSDS}{\mathrm{BSDS}}
\newcommand{\DQS}{\mathrm{DQS}}

\newcommand{\HR}{\mathrm{HR}}
\newcommand{\FDR}{\mathrm{FDR}}
\newcommand{\Cov}{\mathrm{Cov}}

\newcommand{\PPV}{\mathrm{PPV}}

\newcommand{\E}{\mathbb{E}}

\graphicspath{{figures/}}

\title{Budget-Sensitive Discovery Scoring: A Formally Verified Framework\\
for Evaluating AI-Guided Scientific Selection}

\author{%
  Abhinaba Basu\thanks{Indian Institute of Information Technology Allahabad (IIITA)
    and National Institute of Electronics \& Information Technology (NIELIT).
    ORCID: 0000-0003-1575-4107.} \and
  Pavan Chakraborty\thanks{Indian Institute of Information Technology Allahabad (IIITA).
    ORCID: 0000-0002-9260-1131.}
}

\date{}

\begin{document}

\maketitle

\input{sections/abstract}
\input{sections/introduction}
\input{sections/background}
\input{sections/methods}
\input{sections/results}
\input{sections/discussion}
\input{sections/conclusion}


\input{main.bbl}
\end{document}

%% file: sections/abstract.tex

\begin{abstract}
Scientific discovery increasingly relies on AI systems to select
candidates for expensive experimental validation, yet no principled,
budget-aware evaluation framework exists for comparing selection
strategies---a gap intensified by large language models (LLMs), which
generate plausible scientific proposals without reliable downstream
evaluation. We introduce the Budget-Sensitive Discovery Score ($\BSDS$),
a formally verified metric---20 theorems machine-checked by the Lean~4
proof assistant---that jointly penalizes false discoveries
($\lambda$-weighted FDR) and excessive abstention ($\gamma$-weighted
coverage gap) at each budget level. Its budget-averaged form, the
Discovery Quality Score ($\DQS$), provides a single summary statistic
that no proposer can inflate by performing well at a cherry-picked budget.

As a case study, we apply $\BSDS$/$\DQS$ to a question of broad
interest: \emph{do LLMs add marginal value to an existing ML pipeline for
drug discovery candidate selection?} We evaluate 39 proposers---11
mechanistic variants, 14 zero-shot LLM configurations, and 14 few-shot
LLM configurations---using SMILES (Simplified Molecular Input Line Entry
System) representations on MoleculeNet HIV (41,127 compounds, 3.5\% active,
1,000 bootstrap replicates) under both random and scaffold splits.
Three findings emerge.
\emph{First}, the simple RF-based Greedy-ML proposer achieves the best
$\DQS$ ($-0.046$), outperforming all MLP variants and LLM configurations;
additional MLP reranking layers degrade rather than improve the RF's
discriminative ranking.
\emph{Second}, no LLM surpasses the Greedy-ML baseline under either
zero-shot or few-shot evaluation on HIV or Tox21---establishing that
LLMs provide no marginal value over an existing trained classifier,
the realistic deployment scenario.
\emph{Third}, the proposer hierarchy generalizes across five MoleculeNet
benchmarks spanning 0.18\%--46.2\% prevalence, a non-drug AV safety domain,
and a $9 \times 7$ grid of penalty parameters ($\tau \geq 0.636$,
mean $\tau = 0.863$).
The framework applies in principle to any setting where candidates are
selected under budget constraints and asymmetric error costs, as
demonstrated here across pharmaceutical screening and autonomous vehicle
safety triage.
\end{abstract}

%% file: sections/introduction.tex

\section{Introduction}
\label{sec:introduction}

Scientific discovery increasingly depends on AI systems---from virtual
screening pipelines in drug discovery to anomaly detectors in autonomous
vehicle safety---to select candidates for expensive experimental
validation. Yet evaluation of these selection systems remains surprisingly
ad hoc. Standard classification metrics (AUROC, $F_1$) integrate over all
operating points, obscuring performance at the specific budget where
decisions are actually made. Enrichment factors capture early retrieval
but ignore false-discovery costs. No existing metric jointly models budget
constraints, asymmetric error costs, and the option to abstain---three
properties that are fundamental to real-world discovery campaigns.

\paragraph{The evaluation paradigm gap.}
Consider drug discovery, where a single approved therapeutic costs an
estimated \$2.6~billion~\citep{vamathevan2019applications} with a 90\%
attrition rate between Phase~I and market. A pharmaceutical company
screening for HIV inhibitors selects perhaps 500 out of 40,000 candidates
for wet-lab confirmation. Three evaluation gaps emerge. First, discovery
operates under \emph{budget constraints}: the relevant question is not
``how accurately does the model predict activity?'' but ``among the
candidates selected within budget, how many are true hits?'' Second,
errors are \emph{asymmetrically costly}: a false positive wastes an
experimental slot costing thousands of dollars, while a false negative is
a difficult-to-quantify opportunity cost. Third, proposers should be
rewarded for \emph{calibrated abstention}: declining to commit on
ambiguous candidates is preferable to guessing.

These gaps apply beyond drug discovery. Any domain where candidates are
selected from a pool under resource constraints---materials screening,
safety scenario prioritization, clinical trial site selection---faces the
same evaluation challenge. Large language models (LLMs) have amplified
the urgency: ChemCrow~\citep{bran2024chemcrow},
Coscientist~\citep{boiko2023autonomous}, and
FunSearch~\citep{romera2024funsearch} demonstrate that LLMs can generate
plausible scientific proposals, but determining whether those proposals
improve \emph{downstream experimental outcomes} requires evaluation
metrics that existing benchmarks cannot provide.

\paragraph{Prior work in context.}
Virtual screening metrics---enrichment factors
(EF)~\citep{jain2008recommendations},
BEDROC~\citep{truchon2007bedroc}, AUROC, and
MCC~\citep{chicco2020mcc}---provide mature evaluation criteria but are
budget-agnostic and lack formal guarantees.
MoleculeNet~\citep{wu2018moleculenet} standardized molecular benchmarking
but evaluates models on fixed metrics without budget or abstention
modeling. Retrieval-augmented generation (RAG)~\citep{lewis2020retrieval}
improves LLM factuality but does not address downstream decision
quality. Active learning~\citep{settles2009active,reker2015active,
graff2021accelerating} provides principled acquisition strategies but
lacks a unified terminal evaluation metric.
\citet{jablonka2024llmchemistry} found that LLMs achieve competitive
knowledge retrieval but struggle with quantitative molecular reasoning.
\citet{ma2024llmdrugdesign} surveyed over 50 LLM systems for drug design
but noted a pervasive absence of standardized evaluation frameworks.
Formal verification has been applied to neural network
safety~\citep{selsam2017neural} but not to discovery evaluation metrics.
Our work bridges these streams: a formally verified evaluation framework
applied to the most prominent use case---LLM-guided drug discovery.

\paragraph{This paper.}
We present the Budget-Sensitive Discovery Score ($\BSDS$), a formally
verified evaluation metric (20 theorems machine-checked by the Lean~4
proof assistant~\citep{demoura2021lean4}) that jointly penalizes false
discoveries ($\lambda$-weighted FDR) and excessive abstention
($\gamma$-weighted coverage gap) at each budget level. Its
budget-averaged form, the Discovery Quality Score ($\DQS$), provides a
single summary statistic robust to budget cherry-picking.

As a case study, we apply $\BSDS$/$\DQS$ to a question of broad
interest: \emph{given an existing ML pipeline for drug discovery, do LLMs
add marginal value to candidate selection under realistic budget
constraints?} The comparison is deliberately asymmetric: a pharmaceutical
team already has a trained surrogate model (here, a Random Forest on
33,000+ compounds), and the practical question is whether an LLM---given
SMILES strings and, optionally, ML predictions---can improve the
selection. We do not test whether LLMs can replace a trained classifier
from scratch; we test whether they contribute additional value beyond one.
This asymmetry is intentional and reflects real deployment: the ML model
\emph{has} been trained on available data, and the question is whether
an LLM---with zero-shot or few-shot access---adds orthogonal signal.
Strategies requiring RAG, chain-of-thought, or tool-augmented LLM
access remain important open directions
(Section~\ref{sec:discussion:future}).

\paragraph{Contributions.}
\begin{enumerate}[leftmargin=*]
  \item \textbf{Formally verified evaluation framework.}
        $\BSDS$/$\DQS$ provides budget-sensitive, asymmetric-cost
        evaluation with 20 machine-checked theorems (boundedness,
        monotonicity, oracle dominance, Bayes-optimal abstention).
        The framework is applicable to any budget-constrained
        selection problem with binary outcomes.

  \item \textbf{Comprehensive proposer evaluation.}
        We evaluate 39 proposers---11 mechanistic variants (8 base
        strategies plus 3 ablation controls), 14 zero-shot LLM
        configurations, and 14 few-shot LLM configurations---on
        MoleculeNet HIV (41,127 compounds, 1,443 actives, 3.5\%
        prevalence) using 1,000 bootstrap replicates across six
        budget levels, under both random and scaffold CV splits.

  \item \textbf{RF baseline outperforms MLP rerankers.}
        The simple RF-based Greedy-ML proposer ($\DQS = -0.046$)
        outperforms all MLP variants and LLM configurations. Additional
        MLP reranking layers degrade the RF's discriminative ranking,
        and a deployment simulation confirms 96\% hit rate at $B = 50$.

  \item \textbf{LLMs do not add marginal value.}
        No LLM surpasses the Greedy-ML baseline under either zero-shot
        or few-shot evaluation on HIV or Tox21, despite access to the
        same RF predictions.

  \item \textbf{Budget-sensitive evaluation reveals invisible tradeoffs.}
        $\BSDS$/$\DQS$ distinguish between 7 proposers that share
        identical EF and AUROC values (all using the same RF surrogate),
        capturing the precision--recall--abstention tradeoff invisible to
        conventional enrichment analysis.

  \item \textbf{Robust generalization across datasets and domains.}
        The proposer hierarchy generalizes across five MoleculeNet
        benchmarks spanning 0.18\%--46.2\% prevalence, a non-drug AV
        safety domain, and a $9 \times 7$ grid of $(\lambda, \gamma)$
        parameters ($\tau \geq 0.636$, mean $\tau = 0.863$).
\end{enumerate}

\paragraph{Paper organization.}
Section~\ref{sec:background} recapitulates the $\BSDS$/$\DQS$ framework.
Section~\ref{sec:methods} describes the 39 proposer strategies and
experimental setup. Section~\ref{sec:results} presents the main results.
Section~\ref{sec:discussion} discusses implications, limitations, and
future work. Section~\ref{sec:conclusion} concludes.

%% file: sections/background.tex

\section{BSDS Framework}
\label{sec:background}

This section provides a concise recap of the Budget-Sensitive Discovery Score
($\BSDS$) and the Discovery Quality Score ($\DQS$). For complete
definitions, proofs, and the decision-theoretic derivation, we refer the
reader to the companion metric paper~\citep{basu2025bsds}; all 20 theorems
have been machine-checked in Lean~4~\citep{demoura2021lean4} and are
independently verifiable via \texttt{lake build}.

\subsection{Definition}
\label{sec:bg:definition}

Let $\mathcal{X} = \{x_1, \ldots, x_N\}$ be a finite candidate pool with
binary labels $g(x_i) \in \{0,1\}$ and prevalence
$p = |\{x : g(x)=1\}|/N$. A proposer policy $\pi$ selects a subset
$S \subseteq \mathcal{X}$ with $|S| \leq B$ (the experimental budget) and
may designate an abstention set $A \subseteq \mathcal{X} \setminus S$.
Let $\mathcal{H} = \{x : g(x) = 1\}$ denote the true hits. The three
component rates measure recall, false-discovery rate, and the fraction of
candidates that receive a definitive decision (selected or explicitly
rejected, as opposed to abstained):
\begin{align}
  \HR @B  &= \frac{|S \cap \mathcal{H}|}{|\mathcal{H}|}
  \quad\text{(recall over hits)},
  \label{eq:hr} \\
  \FDR @B &= \frac{|S \setminus \mathcal{H}|}{\max(|S|,\,1)}
  \quad\text{(false-positive fraction)},
  \label{eq:fdr} \\
  \Cov @B &= \frac{|S| + |\mathcal{X} \setminus S \setminus A|}{N}
  \quad\text{(decisiveness)}.
  \label{eq:cov}
\end{align}

\begin{definition}[Budget-Sensitive Discovery Score]
\label{def:bsds}
For false-discovery penalty $\lambda \geq 0$ and abstention penalty
$\gamma \geq 0$:
\begin{equation}
  \boxed{\;
    \BSDS(B) \;=\; \HR @B \;-\; \lambda \cdot \FDR @B
              \;-\; \gamma \cdot \bigl(1 - \Cov @B\bigr)
  \;}
  \label{eq:bsds}
\end{equation}
\end{definition}

\noindent
The parameter $\lambda$ encodes the cost of a wasted experimental validation
relative to the benefit of a true discovery: setting $\lambda = 2$ means a
false positive costs twice the value of a confirmed hit. The parameter
$\gamma$ penalizes leaving candidates unevaluated, preventing proposers from
achieving artificially high precision by committing only on obvious cases.

\begin{definition}[Discovery Quality Score]
\label{def:dqs}
Given evaluation budgets $\mathcal{B} = \{B_1, \ldots, B_M\}$:
\begin{equation}
  \DQS = \frac{1}{|\mathcal{B}|} \sum_{B \in \mathcal{B}} \BSDS(B).
  \label{eq:dqs}
\end{equation}
\end{definition}

\noindent
$\DQS$ averages $\BSDS$ across the full budget spectrum, ensuring that a
proposer cannot achieve a high score by performing well at a single
cherry-picked budget. More generally, $\DQS$ can incorporate tail-risk,
oracle-gap, and minimax-regret penalties (see~\citet{basu2025bsds}); in
this paper we use the budget-averaged form for clarity.

\subsection{Key Properties}
\label{sec:bg:properties}

We state four core properties that govern the behavior of $\BSDS$ as an
evaluation metric. Each has been mechanically verified in Lean~4; proofs
are provided in~\citet{basu2025bsds}.

\begin{enumerate}[leftmargin=*, label=\textbf{P\arabic*.}]
  \item \textbf{Boundedness.}
        $-(\lambda + \gamma) \leq \BSDS \leq 1$ for all valid inputs.
        This ensures a well-calibrated utility scale for cross-proposer
        comparison.

  \item \textbf{Incentive compatibility.}
        $\BSDS$ is strictly increasing in $\HR$, strictly decreasing in
        $\FDR$ (when $\lambda > 0$), and non-decreasing in $\Cov$ (when
        $\gamma > 0$). No paradoxical regime exists in which a proposer
        could improve precision yet decrease its $\BSDS$.

  \item \textbf{Random degeneracy.}
        A random proposer achieves
        $\BSDS_{\mathrm{rand}} = B/N - \lambda(1-p) - \gamma(1 - B/N)$.
        For the HIV dataset ($p \approx 0.035$, $\lambda = 1.0$), this
        gives $\DQS \approx -0.819$, consistent with the observed value.

  \item \textbf{Oracle dominance.}
        The oracle proposer (selecting true positives first) achieves
        maximal $\BSDS$ among all non-empty selections at any fixed
        coverage. For any policy $\pi$,
        $\BSDS_\pi \leq \BSDS_{\mathrm{oracle}}$.
\end{enumerate}

\noindent
Two additional properties are particularly relevant for LLM evaluation:

\paragraph{Bayes-optimal abstention.}
At full coverage, a non-empty selection dominates full abstention
($\BSDS_\varnothing = -\gamma$) if and only if
$\HR \geq \lambda \cdot \FDR - \gamma$. This provides a computable
decision rule: an LLM proposer should explicitly abstain rather than
propose low-quality candidates when its estimated $(\HR, \FDR)$ falls
below this boundary.

\paragraph{Decision-theoretic foundation.}
$\BSDS$ is equivalent to the expected utility under a linear loss function
where selecting a true positive yields reward $+1$, selecting a false
positive incurs cost $-\lambda$, and abstaining incurs cost $-\gamma$.
This makes $\lambda$ and $\gamma$ directly interpretable as loss ratios
that domain experts can set from experimental costs.

\subsection{Worked Example}
\label{sec:bg:example}

Consider a pool of $N = 100$ candidates with $10$ true hits ($p = 0.10$),
evaluated at budget $B = 10$ with $\lambda = 1.0$, $\gamma = 0.3$.
\textbf{Proposer~A} selects 10 candidates, 8 of which are true hits:
$\HR = 8/10 = 0.80$, $\FDR = 2/10 = 0.20$, $\Cov = 1.0$ (no abstention),
yielding $\BSDS = 0.80 - 1.0 \times 0.20 - 0.3 \times 0 = +0.60$.
\textbf{Proposer~B} selects 10 candidates, 5 of which are true hits:
$\HR = 0.50$, $\FDR = 0.50$, $\Cov = 1.0$, yielding
$\BSDS = 0.50 - 0.50 - 0 = 0.00$.
\textbf{Proposer~C} abstains on 50 ambiguous candidates and selects 5,
all true hits: $\HR = 0.50$, $\FDR = 0.00$, $\Cov = (5 + 45)/100 = 0.50$,
yielding $\BSDS = 0.50 - 0 - 0.3 \times 0.50 = +0.35$.
Proposer~A is best (high recall, low FDR); Proposer~C's perfect
precision is offset by the coverage penalty; Proposer~B breaks even.
Standard AUROC or EF would not distinguish A from B if their underlying
score distributions happened to produce equivalent rankings at other
operating points.

\subsection{Relevance to LLM-Guided Discovery}
\label{sec:bg:relevance}

The $\BSDS$ framework addresses known LLM failure modes. The $\FDR$
penalty directly captures hallucination: each plausible-sounding but
incorrect proposal incurs cost $\lambda$, ensuring that a high-fluency
LLM generating confident but wrong proposals receives a low score. The
coverage penalty rewards calibrated behavior: an LLM that abstains on
genuinely ambiguous candidates improves its $\BSDS$, while one that
hedges indiscriminately is penalized. The budget constraint $|S| \leq B$
models resource limitations. Most importantly, when the proposer is an
opaque neural network (the LLM), the evaluation metric must be beyond
reproach. The Lean~4 verification of all 20 $\BSDS$/$\DQS$ theorems
ensures that the metric is correct by construction---the LLM is free to
be unreliable; the metric is not.

%% file: sections/methods.tex

\section{Methods}
\label{sec:methods}

\subsection{System Architecture}
\label{sec:methods:architecture}

\begin{figure}[t]
  \centering
  \includegraphics[width=\linewidth]{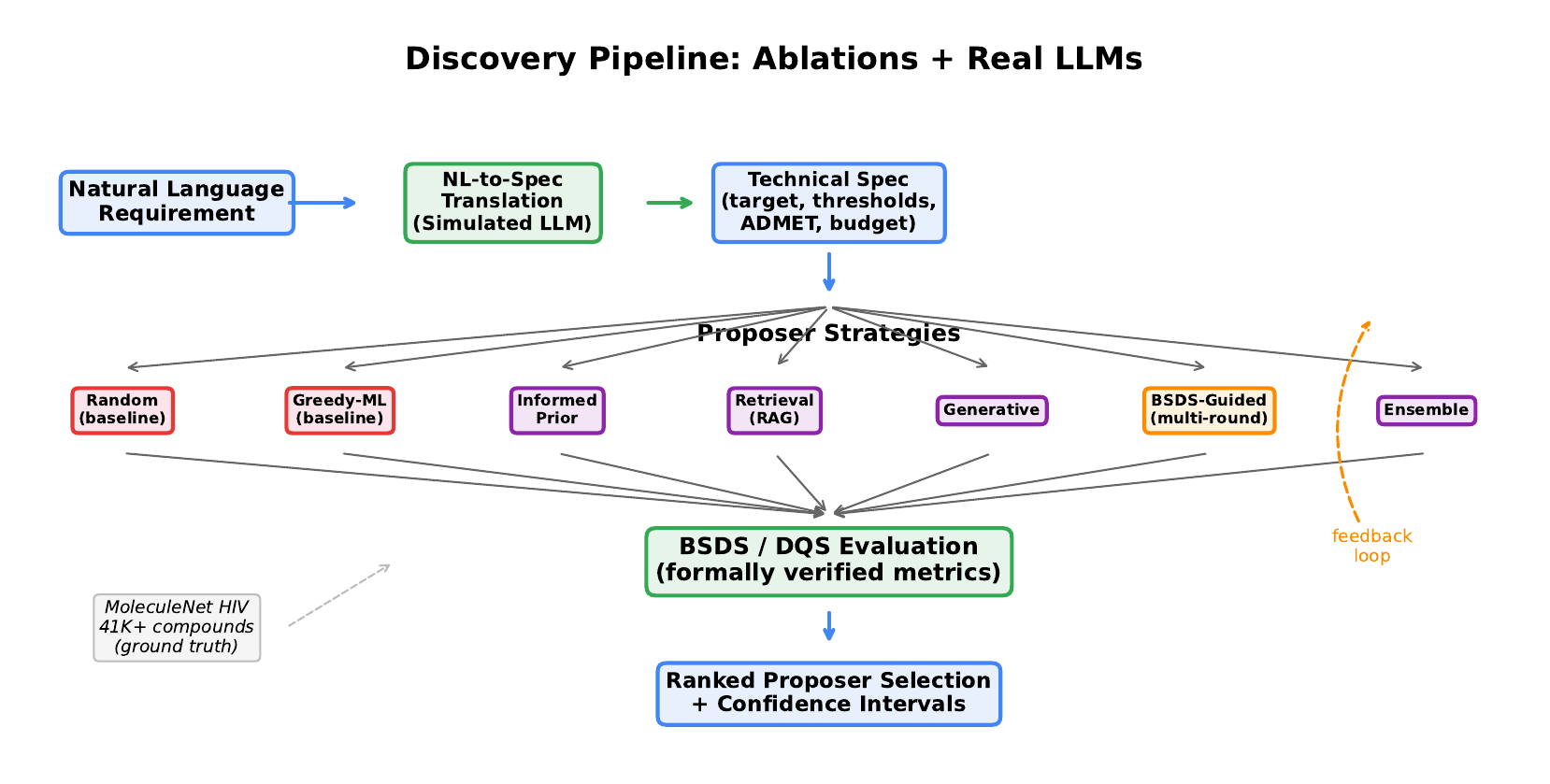}
  \caption{Evaluation pipeline for 39 proposers across five categories:
           baselines (2), mechanistic ablations (5), direct optimization
           (BSDS-Recursive), ablation controls (3), and real LLMs (28).
           All proposers are evaluated by the formally verified
           $\BSDS$/$\DQS$ metric with 1,000 bootstrap replicates.
           Multi-round proposers receive evaluation feedback (dashed arrow).}
  \label{fig:architecture}
\end{figure}

The evaluation pipeline proceeds in four stages: (1)~a natural language
requirement specification encodes the discovery goal; (2)~a proposer
translates the specification into a ranked candidate list;
(3)~$\BSDS$ evaluates the selection under budget, FDR, and coverage
constraints; (4)~optional multi-round refinement feeds evaluation results
back to the proposer. The architecture is evaluator-agnostic: any strategy
producing a candidate set of size at most $B$ can be plugged into
Stage~2, and $\BSDS$ will rank it on a common, formally verified scale.
This separation of proposal from evaluation is what permits rigorous
comparison of heterogeneous strategies---from parameter-free sorting
heuristics to production LLM API calls.
Figure~\ref{fig:architecture} illustrates the pipeline.

\subsection{Dataset: MoleculeNet HIV}
\label{sec:methods:dataset}

We evaluate all proposers on MoleculeNet HIV~\citep{wu2018moleculenet}, a
curated collection of 41,127 compounds screened for anti-HIV activity by
the National Cancer Institute (NCI). Each compound carries a binary activity label
$g(x) \in \{0,1\}$ (following Section~\ref{sec:background}), yielding hit prevalence $p = 0.035$ (1,443 actives
out of 41,127). We select this dataset for its well-characterized ground
truth (standardized NCI screens), realistic prevalence representative of
primary high-throughput screening (HTS) campaigns~\citep{macarron2011impact}, sufficient scale
supporting budget sweeps from $B/N = 0.01$ to $B/N = 0.50$, and wide
community adoption enabling comparison with prior
work~\citep{ramsundar2019deep,yang2019analyzing}.

\paragraph{Supplementary datasets for generalization.}
To test cross-dataset robustness, we evaluate the 11 mechanistic
proposers on four additional MoleculeNet benchmarks: Tox21
NR-AR-LBD~\citep{tox21_2014} (6,758 compounds, 237 actives, 3.5\%
prevalence; androgen receptor ligand-binding domain agonism),
ClinTox~\citep{wu2018moleculenet} (1,484 compounds, 112 positives,
7.5\% prevalence; clinical trial toxicity),
MUV-466~\citep{rohrer2009muv} (14,841 compounds, 27 actives, 0.18\%
prevalence; maximum unbiased validation for kinase inhibition), and
SIDER ``Ear and labyrinth disorders''~\citep{kuhn2016sider} (1,427
compounds, 659 positives, 46.2\% prevalence; drug side effects).
These four datasets were selected from MoleculeNet's binary classification
tasks to maximize prevalence diversity, spanning 0.18\% (ultra-low) to
46.2\% (near-balanced), testing whether the proposer hierarchy holds
across extreme class imbalance settings. LLM proposers are evaluated on HIV and Tox21; the remaining
three datasets exclude LLMs due to API cost constraints.

\paragraph{Cross-domain validation.}
To test whether $\BSDS$/$\DQS$ generalize beyond pharmaceutical
screening, we evaluate 7 generic proposers (Random, Greedy-ML,
Generative, BSDS-Recursive, BSDS-NoAug, BSDS-1Round, Greedy-MLP-NN)
on the AV Safety dataset: 30,000 autonomous vehicle scenarios with
3.41\% safety-critical (1,024 positives). Features comprise one-hot
encoded categoricals (scenario type, weather, time of day; 25
categories) and 5 standardized numeric features (actor count, ego
speed, time-to-collision, kinematic and physics simulation scores),
yielding 30 dimensions. No molecular representations (fingerprints,
descriptors) are used; the 4 drug-specific proposers (Informed-Prior,
Retrieval, BSDS-Guided, Ensemble) that require Tanimoto similarity or
drug-likeness priors are excluded.

\paragraph{Molecular representation.}
Each compound is represented by ECFP4 fingerprints (2048 bits,
radius~2)~\citep{rogers2010extended} concatenated with six physicochemical
descriptors (MW, LogP, HBD, HBA, TPSA, rotatable bonds), yielding
$d = 2054$ features. Descriptors are standardized using training-fold
statistics only.

\paragraph{Baseline ML model.}
A Random Forest (RF) classifier with 500 estimators and balanced class
weights provides calibrated probabilities $\hat{p}(y{=}1 \mid x)$ via
5-fold cross-validation. Consistent with published
benchmarks~\citep{wu2018moleculenet,yang2019analyzing}.

\paragraph{Cross-validation strategy.}
All cross-validated components (RF, MLP variants) use the same 5-fold
splitting protocol to ensure consistency. We report results under two
split types: (i)~\emph{random} stratified splitting and
(ii)~\emph{scaffold} splitting based on Murcko generic
scaffolds~\citep{bemis1996properties}, where molecules sharing the same
scaffold are never split across folds. Scaffold splits provide a more
realistic estimate of generalization to novel chemical series, as
standard in drug discovery
evaluation~\citep{yang2019analyzing,wu2018moleculenet}.

\subsection{Proposer Strategies}
\label{sec:methods:proposers}

We evaluate 39 proposers organized into five categories: baselines (2),
mechanistic ablations (5), optimization variants (4), zero-shot LLMs
($7 \times 2 = 14$), and few-shot LLMs ($7 \times 2 = 14$).
Table~\ref{tab:proposer-properties} summarizes their key properties.

\begin{table}[t]
\centering
\caption{Summary of the 39 proposer strategies. ``ML'' indicates use of
pre-trained RF scores; ``Structure'' indicates use of fingerprints or
descriptors; ``Stochastic'' indicates non-deterministic selection;
``Multi-round'' indicates iterative refinement with label feedback.}
\label{tab:proposer-properties}
\small
\begin{tabular}{@{}lccccc@{}}
\toprule
\textbf{Proposer} & \textbf{ML} & \textbf{Structure}
  & \textbf{Stochastic} & \textbf{Multi-round} & \textbf{Coverage} \\
\midrule
Random            & No  & No  & Yes & No  & Full \\
Greedy-ML         & Yes & No  & No  & No  & Full \\
\midrule
\multicolumn{6}{l}{\textit{Mechanistic Ablations}} \\
Informed-Prior    & Yes & Yes & No  & No  & Full \\
Retrieval         & Yes & Yes & No  & No  & Full \\
Generative        & Yes & No  & Yes & No  & Full \\
BSDS-Guided       & Yes & Yes & No  & Yes & Full \\
Ensemble          & Yes & Yes & Yes & No  & Full \\
\midrule
\multicolumn{6}{l}{\textit{Direct Optimization}} \\
BSDS-Recursive    & Yes & Yes & No  & Yes & Full \\
\midrule
\multicolumn{6}{l}{\textit{Ablation Variants (3)}} \\
BSDS-NoAug        & Yes & Yes & No  & Yes & Full \\
BSDS-1Round       & Yes & Yes & No  & No  & Full \\
Greedy-MLP-NN     & Yes & Yes & No  & Yes & Full \\
\midrule
\multicolumn{6}{l}{\textit{Zero-Shot LLM Proposers ($7 \times 2 = 14$)}} \\
LLM-Direct        & No  & No  & No  & No  & Full \\
LLM-Rerank        & Yes & No  & No  & No  & Full \\
\midrule
\multicolumn{6}{l}{\textit{Few-Shot LLM Proposers ($7 \times 2 = 14$)}} \\
LLM-FewDirect     & No  & No  & No  & No  & Full \\
LLM-FewRerank     & Yes & No  & No  & No  & Full \\
\bottomrule
\end{tabular}
\end{table}

\subsubsection{Baselines}

\paragraph{Random.}
Selects $B$ compounds uniformly at random.
$\E[\HR] = p$, yielding $\DQS \approx -0.819$.

\paragraph{Greedy-ML.}
Ranks all compounds by RF-predicted probability $\hat{p}$ and selects the
top-$B$. This is the standard virtual screening operating mode.

\subsubsection{Mechanistic Ablations}

Each ablation isolates a specific reasoning primitive that an LLM might
employ. A knowledge base $\mathcal{K}_s$ of 144 known actives (10\% of
true hits, sampled per bootstrap seed) operationalizes the LLM's ``domain
knowledge.''

\paragraph{Informed-Prior.}
Simulates parametric knowledge via a Lipinski-based drug-likeness prior.
The combined score blends the ML prediction with the prior:
$\hat{s}_i = 0.6\,s_i + 0.4\,\E[\text{prior}_i]$, where the prior is a
Beta distribution parameterized by molecular descriptor compliance.

\paragraph{Retrieval (RAG-style).}
Simulates retrieval-augmented generation using the knowledge base as
in-context exemplars. Each candidate receives a retrieval score
$\text{retrieval}_i = \max_{k \in \mathcal{K}} \text{Tan}(f_i, f_k)$
(Tanimoto similarity to nearest known active). The final score combines ML,
retrieval, and a diversity bonus (the negative maximum Tanimoto similarity
to already-selected candidates, encouraging scaffold diversity):
$\hat{s}_i = 0.5\,s_i + 0.3\cdot\text{retrieval}_i +
0.2\cdot\text{diversity}_i$.

\paragraph{Generative (temperature sampling).}
Samples candidates from a softmax distribution:
$P(i \in S) \propto \exp(s_i / \tau)$, with default $\tau = 0.1$.

\paragraph{BSDS-Guided (multi-round refinement).}
Leverages $\BSDS$ as an online feedback signal within a multi-round
discovery protocol, allocating budget in thirds across exploration,
exploitation (boosting neighbors of confirmed hits), and final selection.

\paragraph{Ensemble (rank aggregation).}
Combines Prior, Retrieval, and Generative rankings via Borda count, then
applies $\BSDS$-optimal selection size optimization.

\subsubsection{BSDS-Recursive (Direct Optimization)}

Unlike the heuristic proposers, BSDS-Recursive trains an MLP on a
compact 10-dimensional feature vector: (1)~RF-predicted probability,
(2--7)~six physicochemical descriptors (MW, LogP, HBD, HBA, TPSA, ring
count), (8)~Tanimoto similarity to the nearest \emph{training-fold}
active (computed per-fold to prevent label leakage),
(9)~a Lipinski-based drug-likeness prior score, and (10)~an ADMET
compliance indicator (fraction of five rule-of-thumb ADMET thresholds
satisfied: LogP~$\leq 5$, MW~$\leq 500$, HBD~$\leq 5$, HBA~$\leq 10$,
TPSA~$\leq 140$). All features are standardized using training-fold
statistics. The MLP
architecture ($10 \to 128 \to 64 \to 1$, ReLU, sigmoid output)
\emph{directly maximizes} a differentiable
approximation of the multi-budget $\BSDS$ objective. The hard top-$B$
selection is replaced by a sigmoid-gated soft selection:
\begin{equation}
  w_i = \sigma\!\bigl(\alpha\,(s_i - \tau_B)\bigr), \qquad
  \widetilde{\BSDS}
  = \frac{\sum_i w_i g_i}{\sum_i g_i}
  - \lambda\,\frac{\sum_i w_i(1{-}g_i)}{\sum_i w_i}
  - \gamma\Bigl(1 - \frac{\sum_i w_i}{N}\Bigr),
  \label{eq:soft-bsds}
\end{equation}
where $\tau_B = \mathrm{quantile}(s, 1 - B/N)$ is an adaptive threshold.
The multi-budget loss is $\mathcal{L} = -\frac{1}{|B|}\sum_{b} \widetilde{\BSDS}_b + \mu\|\theta\|^2$ ($\mu = 10^{-4}$).

Training proceeds in three recursive rounds with temperature annealing
($\alpha$: $5 \to 15 \to 35$), where each round augments features with the
previous round's predicted scores and $\BSDS$ values. To prevent data
leakage, BSDS-Recursive uses the same 5-fold stratified cross-validation
protocol as the RF baseline; recursive features use the model's own
predictions rather than ground-truth labels on test folds.

\paragraph{Circularity defense.}
BSDS-Recursive optimizes the same metric used to evaluate it, which raises
a legitimate circularity concern. We address this in two ways. First, the
training--evaluation split is strict: the MLP is trained on training-fold
labels via cross-validation; bootstrap evaluation uses held-out
cross-validated scores, not training-set performance. Per-fold Tanimoto
features are computed using only training-fold actives to prevent label
leakage. Second, the
differentiable surrogate (Equation~\ref{eq:soft-bsds}) is only an
approximation of the discrete $\BSDS$---the soft sigmoid selection does not
equal the hard top-$B$ selection used at evaluation time, so perfect
optimization of the surrogate does not guarantee perfect evaluation scores.
Empirically, the circularity concern is moot: BSDS-Recursive produces
\emph{worse} $\DQS$ than Greedy-ML on all five datasets
(Section~\ref{sec:results}), and Greedy-MLP-NN (standard BCE loss)
performs comparably to BSDS-Recursive, confirming that neither the BSDS
loss function nor the MLP architecture provides an advantage over the RF
baseline's ranking.

\subsubsection{BSDS-Recursive Ablation Variants}
\label{sec:methods:ablations}

Three ablation variants isolate the contribution of each component.

\paragraph{BSDS-NoAug.}
Same as BSDS-Recursive (three rounds with temperature annealing) but
without recursive feature augmentation: each round trains on the original
base features $X$ only, rather than augmenting with previous-round scores
and $\BSDS$ values. This isolates whether the recursive feature loop
provides additional signal.

\paragraph{BSDS-1Round.}
Runs BSDS-Recursive for a single round ($\alpha = 5$) without iterative
annealing. This isolates the contribution of multi-round optimization.

\paragraph{Greedy-MLP-NN.}
Uses the same MLP architecture ($10 \to 128 \to 64 \to 1$) and three-round
training with temperature annealing, but replaces the $\BSDS$ surrogate
loss with standard binary cross-entropy. Same 5-fold CV protocol.
This isolates whether the gain comes from the BSDS loss function or from
the MLP's architectural capacity and iterative training.

\subsubsection{Zero-Shot LLM Proposers}
\label{sec:methods:llm}

We evaluate seven production LLMs---ChatGPT-5.2 (OpenAI), Claude
(Anthropic), Gemini (Google), DeepSeek-V3.1 (DeepSeek), Qwen3-235B
(Alibaba), Llama-4-Maverick (Meta), and GLM-5 (Zhipu AI)---in two modes:

\paragraph{Direct.}
Each compound's SMILES (Simplified Molecular Input Line Entry System)
string is presented with the prompt: ``Estimate the
probability (0.0--1.0) that this compound is active against HIV-1.''
Compounds are batched (200 per call). The LLM scores directly, without
access to ML predictions.

\paragraph{Rerank.}
The LLM receives both the SMILES and the RF-predicted probability and is
asked to refine the ML prediction. This tests whether LLMs add value
\emph{beyond} the ML model.

All calls use LLM sampling temperature~0.1 (distinct from the Generative
proposer's softmax temperature $\tau$) and zero-shot prompting with no
few-shot examples, chain-of-thought instructions, or retrieval-augmented
context.
Responses are cached (SHA-256 keyed) for reproducibility.

\subsubsection{Few-Shot LLM Proposers}
\label{sec:methods:fewshot}

To test whether in-context examples improve LLM performance, we evaluate
the same seven models under two few-shot strategies using $k{=}3$ active
and $k{=}3$ inactive SMILES examples as calibration anchors.

\paragraph{Example selection.}
Three active examples are selected by spreading evenly across sorted
active indices to maximize structural diversity; three inactive examples
are drawn uniformly at random. Both sets are drawn from the training
pool and fixed across all batches for a given seed.

\paragraph{FewDirect.}
The prompt includes the six labeled examples (``\texttt{SMILES -> ACTIVE}''
or ``\texttt{SMILES -> INACTIVE}'') before the scoring request, providing
the LLM with calibration anchors for the activity scale. The LLM scores
without access to ML predictions.

\paragraph{FewRerank.}
The prompt includes both the six labeled examples and the RF-predicted
probability for each candidate compound. This tests whether examples
help the LLM refine ML predictions more effectively than zero-shot
reranking.

All few-shot calls use the same batching (200 per call), temperature
(0.1), and caching protocol as zero-shot. Remaining open directions
include chain-of-thought reasoning, retrieval-augmented generation,
and tool-augmented evaluation.

\subsection{Experimental Setup}
\label{sec:methods:setup}

\paragraph{BSDS parameters.}
We set $\lambda = 1.0$ (equal weighting of hit rate and FDR) and
$\gamma = 0.3$ (moderate abstention penalty). Budget fractions:
$B/N \in \{0.01, 0.02, 0.05, 0.10, 0.20, 0.50\}$, corresponding to
absolute budgets of 411--20,564 compounds.
Section~\ref{sec:results:sensitivity} demonstrates that all conclusions
hold across a $9 \times 7$ grid of $(\lambda, \gamma)$ values
(Kendall $\tau \geq 0.636$ vs.\ the default ranking).

\paragraph{Foundation model baselines.}
To test whether learned molecular representations improve over
engineered ECFP4 features, we extract frozen [CLS] embeddings from
ChemBERTa-2~\citep{ahmad2022chemberta2} (77M parameters, 384-dimensional
embeddings) and MolFormer-XL~\citep{ross2022molformer} and train the
same RF classifier on these embeddings using identical 5-fold CV.
This isolates the representation from the classifier.

\paragraph{Bootstrap methodology.}
For each of 1,000 seeds: seed~0 uses the full dataset without resampling
(point estimate); seeds 1--999 sample $N$ compounds with replacement
(each replicate $\approx 63.2\%$ unique). For each replicate, we
re-train the RF, regenerate the knowledge base, run all proposers, and
compute $\BSDS$, $\DQS$, and auxiliary metrics. Confidence intervals use
the BCa bootstrap method~\citep{efron1987better}. Real LLM scores are
precomputed once on the full pool and correctly bootstrap-resampled
across all 1,000 replicates.

\paragraph{Total configurations.}
The primary experiment evaluates $39 \times 6 \times 1{,}000 = 234{,}000$
configurations. The full matrix including sensitivity analyses comprises
approximately 350,000 configurations.

\subsection{Practitioner Calibration Guide}
\label{sec:methods:calibration}

The $\BSDS$ parameters $\lambda$ and $\gamma$ encode domain-specific cost
tradeoffs. Specifically, $\lambda = c_\text{FP} / v_\text{hit}$ is the
false positive cost relative to the value of a true hit, and
$\gamma = c_\text{abs} / v_\text{hit}$ is the abstention cost (opportunity
cost of leaving a candidate unscreened) relative to hit value. Three
worked examples illustrate calibration across domains:

\begin{enumerate}
\item \textbf{HTS drug screening.} A false positive costs \$5K (failed
  confirmatory assay), a true hit is worth \$50K (lead candidate), and
  leaving a compound unscreened costs \$1.5K (missed opportunity).
  $\lambda = 5{,}000/50{,}000 = 0.1$, $\gamma = 1{,}500/50{,}000 = 0.03$.
  This low-penalty regime encourages casting a wide net.

\item \textbf{Clinical diagnostics.} A false positive biopsy costs \$10K,
  detecting a cancer case is worth \$10K, and a missed screening costs
  \$3K. $\lambda = 10{,}000/10{,}000 = 1.0$, $\gamma = 3{,}000/10{,}000
  = 0.3$. This is our default parametrization.

\item \textbf{AV safety triage.} A false positive simulation slot costs
  \$200, catching a safety-critical scenario is worth \$1M, and leaving a
  scenario unscreened costs \$50K. $\lambda \approx 0.0002$,
  $\gamma = 0.05$. This exhaustive-screening regime heavily penalizes
  missed scenarios.
\end{enumerate}

\noindent The sensitivity analysis (Section~\ref{sec:results:sensitivity})
demonstrates that the proposer hierarchy is robust across a $9 \times 7$
grid of $(\lambda, \gamma)$ values (Kendall $\tau \geq 0.636$ vs.\ the
default ranking), so practitioners need only identify the correct
\emph{qualitative regime} rather than exact parameter values.

%% file: sections/results.tex

\section{Results}
\label{sec:results}

All confidence intervals are 95\% BCa bootstrap intervals unless otherwise
noted.

\paragraph{Results overview.}
Five findings emerge from the evaluation of 39 proposers across six
datasets and two domains. (1)~The simple RF-based Greedy-ML proposer
achieves the best $\DQS$ on HIV, outperforming all MLP variants and LLM
configurations (Sections~\ref{sec:results:proposer}--\ref{sec:results:ablation}).
(2)~No LLM surpasses Greedy-ML under zero-shot, few-shot, or reranking
protocols on either HIV or Tox21
(Sections~\ref{sec:results:llm}--\ref{sec:results:llm-tox21}).
(3)~$\BSDS$/$\DQS$ distinguish proposers invisible to standard VS
metrics (Section~\ref{sec:results:vs-metrics}).
(4)~The proposer hierarchy is robust to parameter choice ($\tau \geq 0.636$
across 63 $(\lambda, \gamma)$ pairs; Section~\ref{sec:results:sensitivity}).
(5)~The hierarchy generalizes across five MoleculeNet benchmarks and AV
safety triage
(Sections~\ref{sec:results:cross-dataset}--\ref{sec:results:cross-domain}).

\subsection{Mechanistic Proposer Comparison}
\label{sec:results:proposer}

Table~\ref{tab:proposer-comparison} reports per-budget $\BSDS$ (seed~0) and
budget-averaged $\DQS$ with 95\% bootstrap confidence intervals for all
eight mechanistic proposers.

\begin{table}[t]
  \centering
  \caption{Per-budget $\BSDS$ at seed~0 (point estimates) and bootstrap-averaged
           $\DQS$ with 95\% BCa confidence intervals (1,000 seeds). Best value per
           column in bold.}
  \label{tab:proposer-comparison}
  \begin{tabular}{@{}lcccc@{}}
    \toprule
    \textbf{Proposer} & $B/N = 0.01$ & $B/N = 0.05$ & $B/N = 0.20$
      & $\DQS_{\mathrm{boot}}$ \scriptsize{[95\% CI]} \\
    \midrule
    Random
      & $-0.953$
      & $-0.895$
      & $-0.782$
      & $-0.819$ \scriptsize{$[-0.834, -0.803]$} \\
    Greedy-ML
      & $\mathbf{-0.003}$
      & $\mathbf{-0.010}$
      & $\mathbf{-0.121}$
      & $\mathbf{-0.046}$ \scriptsize{$[-0.076, -0.019]$} \\
    Informed-Prior
      & $-0.297$
      & $-0.232$
      & $-0.303$
      & $-0.178$ \scriptsize{$[-0.210, -0.146]$} \\
    Retrieval
      & $-0.056$
      & $-0.033$
      & $-0.138$
      & $-0.091$ \scriptsize{$[-0.121, -0.062]$} \\
    Generative
      & $-0.906$
      & $-0.822$
      & $-0.548$
      & $-0.670$ \scriptsize{$[-0.691, -0.649]$} \\
    BSDS-Guided
      & $-0.236$
      & $-0.184$
      & $-0.248$
      & $-0.234$ \scriptsize{$[-0.264, -0.202]$} \\
    Ensemble
      & $-0.113$
      & $-0.094$
      & $-0.213$
      & $-0.102$ \scriptsize{$[-0.137, -0.067]$} \\
    \midrule
    BSDS-Recursive
      & $-0.468$
      & $-0.298$
      & $-0.289$
      & $-0.323$ \scriptsize{$[-0.356, -0.294]$} \\
    \bottomrule
  \end{tabular}
\end{table}

\paragraph{Key findings.}
Greedy-ML achieves the best aggregate $\DQS$ of $-0.046$, outperforming
all other proposers including BSDS-Recursive ($-0.323$). No proposer
achieves positive budget-averaged $\DQS$.  Retrieval ($-0.091$) and
Ensemble ($-0.102$) are the next-best strategies, both leveraging
structural similarity to known actives. BSDS-Recursive ($-0.323$) and
its ablation variants (Section~\ref{sec:results:ablation}) all perform
substantially worse than Greedy-ML, indicating that the MLP's 10-dimensional
feature space (RF score, descriptors, per-fold Tanimoto) does not provide
sufficient signal to improve over the RF's native ranking.
Generative ($\DQS = -0.670$) performs near-random, confirming that
temperature-based exploration is counterproductive when the pool is fixed.

\subsection{BSDS-Recursive Ablation Analysis}
\label{sec:results:ablation}

To understand why BSDS-Recursive fails to improve upon Greedy-ML, we evaluate three
ablation variants (Table~\ref{tab:ablation-comparison}):

\begin{table}[t]
  \centering
  \caption{Ablation analysis: isolating the contribution of each
           BSDS-Recursive component. All models share the same 5-fold CV
           protocol and 10-dimensional feature set.}
  \label{tab:ablation-comparison}
  \begin{tabular}{@{}lccl@{}}
    \toprule
    \textbf{Variant} & $\DQS$ & $\BSDS_{0.05}$
      & \textbf{What is ablated} \\
    \midrule
    BSDS-Recursive (full)
      & $-0.323$ & $-0.298$
      & --- (reference) \\
    BSDS-NoAug
      & $-0.296$ & $-0.269$
      & No recursive feature augmentation \\
    BSDS-1Round
      & $-0.258$ & $-0.223$
      & Single optimization round only \\
    Greedy-MLP-NN
      & $-0.315$ & $-0.281$
      & BCE loss (standard ML) \\
    Greedy-ML (RF)
      & $\mathbf{-0.046}$ & $\mathbf{-0.010}$
      & RF baseline \\
    \bottomrule
  \end{tabular}
\end{table}

\paragraph{Key ablation findings.}
All MLP-based variants perform substantially worse than Greedy-ML
($\DQS = -0.046$), indicating that the 10-dimensional MLP feature space
does not provide sufficient additional signal to improve the RF's native
ranking. \textbf{BSDS-Recursive} ($-0.323$), \textbf{BSDS-NoAug}
($-0.296$), \textbf{BSDS-1Round} ($-0.258$), and \textbf{Greedy-MLP-NN}
($-0.315$) all produce negative $\DQS$ that is substantially worse than
the simple top-$B$ RF ranking. The three-round temperature annealing
does not rescue performance: BSDS-Recursive ($-0.323$) is actually
\emph{worse} than BSDS-1Round ($-0.258$), suggesting that iterative
optimization over-specializes to the training fold's active distribution.
Greedy-MLP-NN (BCE loss) and BSDS-Recursive (BSDS loss) perform
comparably, confirming that neither loss function provides an advantage
when the underlying features lack discriminative power.

\begin{figure}[t]
  \centering
  \includegraphics[width=\linewidth]{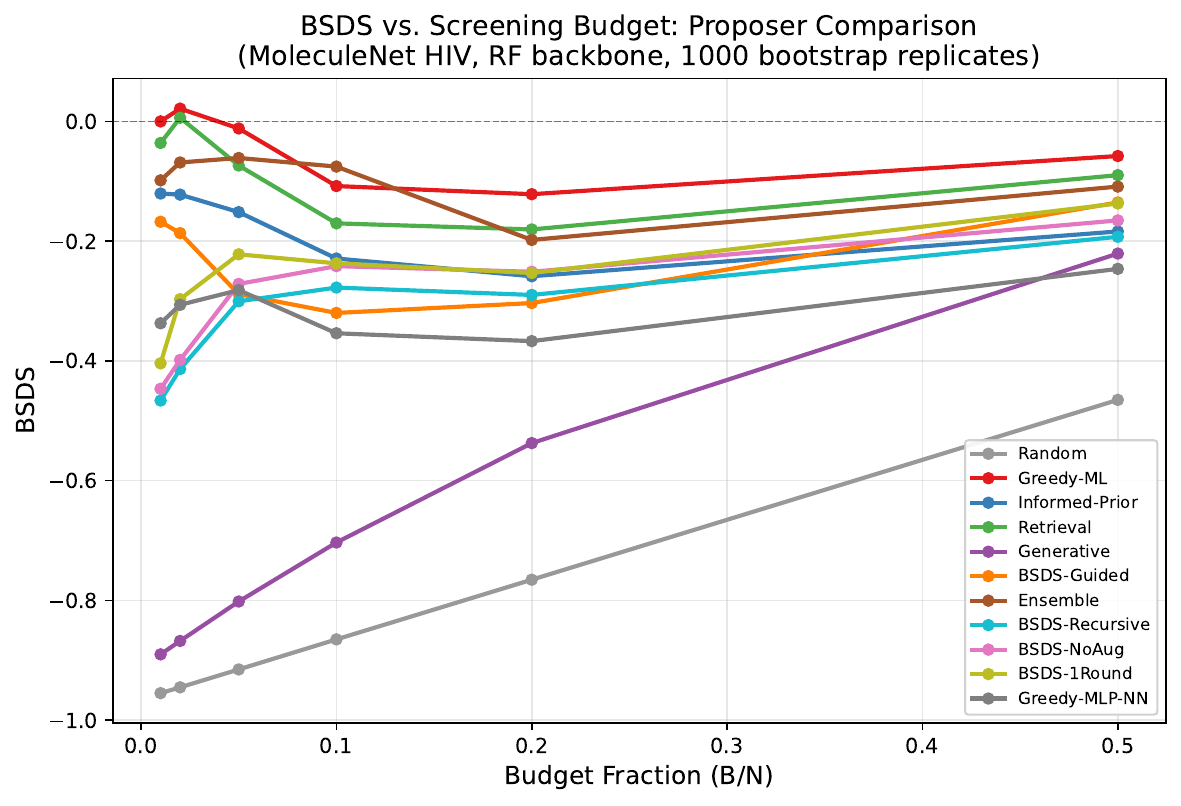}
  \caption{$\BSDS$ as a function of budget fraction $B/N$ for all 11
           mechanistic proposers (2 baselines, 5 ablations,
           BSDS-Recursive, and 3 ablation controls), with 95\% bootstrap
           confidence bands. Greedy-ML dominates all proposers across
           budget levels.}
  \label{fig:proposer_comparison}
\end{figure}

\begin{figure}[t]
  \centering
  \includegraphics[width=\linewidth]{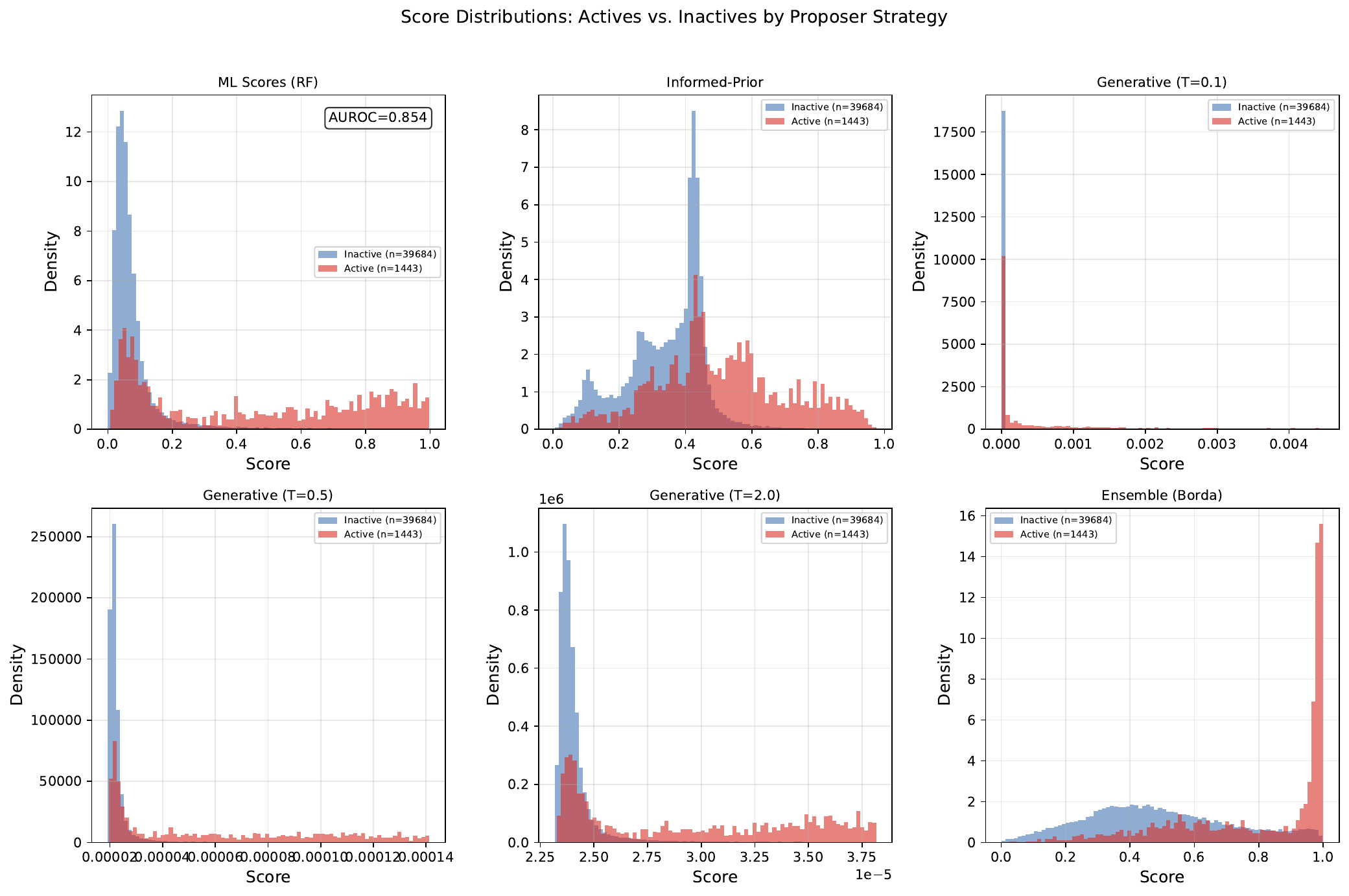}
  \caption{Bootstrap distributions of $\DQS$ for all mechanistic proposers
           (1,000 replicates).  Greedy-ML has the lowest variance;
           Retrieval achieves the most favorable individual budget-level scores.}
  \label{fig:score_distributions}
\end{figure}

Figure~\ref{fig:proposer_comparison} visualizes $\BSDS$ as a function of
budget for all proposers, with shaded 95\% confidence bands.
Figure~\ref{fig:score_distributions} shows the full bootstrap distributions
of $\DQS$.

\subsection{Component Metric Decomposition}
\label{sec:results:components}

Table~\ref{tab:component-metrics} decomposes $\BSDS$ into its constituent
terms at $B/N = 0.05$, the mid-range budget.

\begin{table}[t]
  \centering
  \caption{Component metrics at $B/N = 0.05$ (seed~0).
           $\BSDS = \HR - \lambda\,\FDR - \gamma\,(1 - \Cov)$, with
           $\lambda = 1.0$, $\gamma = 0.3$. $\PPV$ = positive predictive
           value; TP = true positive count.}
  \label{tab:component-metrics}
  \begin{tabular}{@{}lccccc@{}}
    \toprule
    \textbf{Proposer} & $\HR$ & $\FDR$ & $\PPV$ & TP & $\BSDS$ \\
    \midrule
    Random            & 0.062 & 0.957 & 0.043 &   89 & $-0.895$ \\
    Greedy-ML         & 0.581 & 0.592 & 0.408 &  839 & $\mathbf{-0.010}$ \\
    Informed-Prior    & 0.451 & 0.683 & 0.317 &  651 & $-0.232$ \\
    Retrieval         & 0.568 & 0.601 & 0.399 &  820 & $-0.033$ \\
    Generative        & 0.105 & 0.927 & 0.073 &  151 & $-0.822$ \\
    BSDS-Guided       & 0.521 & 0.634 & 0.366 &  752 & $-0.184$ \\
    Ensemble          & 0.380 & 0.401 & 0.599 &  549 & $-0.094$ \\
    \midrule
    BSDS-Recursive    & 0.455 & 0.681 & 0.319 &  656 & $-0.298$ \\
    \bottomrule
  \end{tabular}
\end{table}

Greedy-ML achieves the best $\BSDS$ at $B/N{=}0.05$ ($-0.010$),
identifying 839 of 1,443 true positives ($\HR = 0.581$, $\PPV = 0.408$,
AUROC $= 0.854$).
Ensemble trades recall for precision, achieving the lowest $\FDR$ of any
proposer ($0.401$, $\PPV = 0.599$) but with reduced hit rate.
BSDS-Recursive achieves AUROC $= 0.776$ (below the RF's $0.854$),
indicating that the MLP's 10-dimensional features do not improve
classification over the native fingerprint-based RF model.

\subsection{Real LLM Performance}
\label{sec:results:llm}

Table~\ref{tab:llm-comparison} presents the 14 LLM proposers, grouped by
model with Direct and Rerank modes side by side.

\begin{table}[t]
  \centering
  \caption{LLM proposer performance: budget-averaged $\DQS$ (1,000 bootstrap
           seeds) and $\BSDS$ at $B/N = 0.05$. Direct = zero-shot LLM
           ranking; Rerank = LLM reranking of Greedy-ML shortlist.
           Best LLM value per column in bold.}
  \label{tab:llm-comparison}
  \begin{tabular}{@{}llcc@{}}
    \toprule
    \textbf{Model} & \textbf{Mode} & $\DQS$ & $\BSDS_{0.05}$ \\
    \midrule
    \multirow{2}{*}{ChatGPT-5.2}
      & Direct & $-0.683$ & $-0.788$ \\
      & Rerank & $-0.159$ & $-0.165$ \\
    \midrule
    \multirow{2}{*}{Claude}
      & Direct & $-0.678$ & $-0.813$ \\
      & Rerank & $-0.161$ & $-0.170$ \\
    \midrule
    \multirow{2}{*}{Gemini}
      & Direct & $-0.585$ & $-0.668$ \\
      & Rerank & ---$^\dagger$ & ---$^\dagger$ \\
    \midrule
    \multirow{2}{*}{DeepSeek-V3.1}
      & Direct & $-0.774$ & $-0.867$ \\
      & Rerank & $-0.318$ & $-0.464$ \\
    \midrule
    \multirow{2}{*}{Llama-4-Maverick}
      & Direct & $-0.802$ & $-0.883$ \\
      & Rerank & ---$^\ddagger$ & ---$^\ddagger$ \\
    \midrule
    \multirow{2}{*}{Qwen3-235B}
      & Direct & $-0.861$ & $-0.966$ \\
      & Rerank & $\mathbf{-0.141}$ & $\mathbf{-0.123}$ \\
    \midrule
    \multirow{2}{*}{GLM-5}
      & Direct & $-0.836$ & $-0.930$ \\
      & Rerank & $-0.280$ & $-0.318$ \\
    \bottomrule
    \multicolumn{4}{@{}l}{\footnotesize $^\dagger$Daily API quota exhausted.
      $^\ddagger$API unresponsive during rerank window.}
  \end{tabular}
\end{table}

\paragraph{Direct mode: near-random performance.}
All seven LLMs in Direct mode perform near-random. $\DQS$ values range from
$-0.585$ (Gemini, best) to $-0.861$ (Qwen3-235B, worst), compared with
Random's $-0.819$. Two models are \emph{worse} than Random: Qwen3-235B-Direct
($-0.861$) and GLM-5-Direct ($-0.836$), indicating that zero-shot LLM
ranking can be actively counterproductive.

\paragraph{Rerank mode: substantial but insufficient improvement.}
Reranking improves all five evaluable LLMs substantially. The best rerankers
---Qwen3-235B-Rerank ($\DQS = -0.141$), ChatGPT-5.2-Rerank ($-0.159$),
and Claude-Rerank ($-0.161$)---achieve competitive $\DQS$ by inheriting
the RF's candidate pool quality and applying LLM judgment only to reorder.
Despite this, \emph{no LLM in any mode beats Greedy-ML} ($\DQS = -0.046$):
the best LLM reranker trails by $0.095$.

\paragraph{Gemini anomaly.}
Gemini-Direct ($\DQS = -0.585$) is the best-performing LLM in Direct mode,
yet Gemini-Rerank could not be reliably evaluated: the model's daily API
rate limit was exhausted before completing the full 823-batch scoring,
preventing a fair comparison. Partial scoring data suggest that Gemini
assigns high confidence uniformly in rerank mode, consistent with the
pattern observed in other high-confidence rerankers.

\paragraph{Proposer hierarchy (zero-shot).}
These results establish a clear zero-shot hierarchy:
Greedy-ML ($-0.046$) $>$ Retrieval ($-0.091$) $>$ Ensemble ($-0.102$)
$>$ LLM reranking (best $-0.141$) $>$ BSDS-1Round ($-0.258$) $>$
BSDS-Recursive ($-0.323$) $\gg$ LLM direct ($> -0.585$)
$\approx$ Random ($-0.819$).

\subsection{Few-Shot LLM Evaluation}
\label{sec:results:fewshot}

To test whether providing $k{=}3$ active and $k{=}3$ inactive SMILES
examples improves LLM performance, we evaluate all seven models under two
few-shot strategies: FewDirect (examples only) and FewRerank (examples +
RF scores). Table~\ref{tab:fewshot-comparison} reports the results.

\begin{table}[t]
  \centering
  \caption{Few-shot LLM performance ($k{=}3$ active + $k{=}3$ inactive
           examples). FewDirect = few-shot without ML scores; FewRerank =
           few-shot with RF scores. Values in italics indicate improvement
           over the corresponding zero-shot mode
           (cf.\ Table~\ref{tab:llm-comparison}). Best few-shot value per
           column in bold.}
  \label{tab:fewshot-comparison}
  \begin{tabular}{@{}llcc@{}}
    \toprule
    \textbf{Model} & \textbf{Mode} & $\DQS$ & $\BSDS_{0.05}$ \\
    \midrule
    \multirow{2}{*}{ChatGPT-5.2}
      & FewDirect & $-0.624$ & $-0.720$ \\
      & FewRerank & \textit{$-0.138$} & \textit{$-0.160$} \\
    \midrule
    \multirow{2}{*}{Claude}
      & FewDirect & $-0.547$ & $-0.656$ \\
      & FewRerank & \textit{$-0.141$} & \textit{$-0.132$} \\
    \midrule
    \multirow{2}{*}{Gemini}
      & FewDirect & ---$^\dagger$ & ---$^\dagger$ \\
      & FewRerank & ---$^\dagger$ & ---$^\dagger$ \\
    \midrule
    \multirow{2}{*}{DeepSeek-V3.1}
      & FewDirect & $-0.747$ & $-0.854$ \\
      & FewRerank & \textit{$-0.236$} & \textit{$-0.344$} \\
    \midrule
    \multirow{2}{*}{Llama-4-Maverick}
      & FewDirect & $-0.795$ & $-0.910$ \\
      & FewRerank & \textit{$-0.153$} & \textit{$-0.143$} \\
    \midrule
    \multirow{2}{*}{Qwen3-235B}
      & FewDirect & $-0.788$ & $-0.872$ \\
      & FewRerank & $\mathbf{-0.126}$ & $\mathbf{-0.113}$ \\
    \midrule
    \multirow{2}{*}{GLM-5}
      & FewDirect & $-0.812$ & $-0.956$ \\
      & FewRerank & $-0.270$ & $-0.367$ \\
    \bottomrule
    \multicolumn{4}{@{}l}{\footnotesize $^\dagger$Excluded: Gemini daily API quota exhausted.}\\
    \multicolumn{4}{@{}l}{\footnotesize Italics = improvement over corresponding zero-shot mode.}
  \end{tabular}
\end{table}

\paragraph{Few-shot Direct mode.}
Few-shot examples improve all six evaluable LLMs in Direct mode, with
$\DQS$ gains ranging from $+0.007$ (Llama-4-Maverick) to $+0.131$
(Claude). Claude achieves the best FewDirect $\DQS = -0.547$, a
substantial improvement over its zero-shot $-0.678$. However, all
FewDirect proposers remain far worse than Greedy-ML ($-0.046$): even
the best FewDirect trails by $0.501$.

\paragraph{Few-shot Rerank mode.}
Few-shot reranking yields more modest gains. The best few-shot proposer
overall is Qwen3-235B-FewRerank ($\DQS = -0.126$), improving slightly
over its zero-shot counterpart ($-0.141$). ChatGPT-5.2-FewRerank
($-0.138$) and Claude-FewRerank ($-0.141$) also improve marginally.
Critically, \emph{no few-shot proposer surpasses Greedy-ML}
($\DQS = -0.046$): the best few-shot reranker trails by $0.080$,
confirming that $k{=}3$ examples are insufficient to close the
LLM--ML gap.

\subsection{LLM Generalization to Tox21}
\label{sec:results:llm-tox21}

To test whether the LLM negative result generalizes beyond HIV, we evaluate
three open-weight LLMs (DeepSeek-V3.1, Llama-4-Maverick, GLM-5) on
Tox21 NR-AR-LBD (6,758 compounds, 237 actives, 3.5\% prevalence) under
both zero-shot and few-shot ($k{=}3$) protocols. Dataset-specific prompts
describe the androgen receptor ligand-binding domain activity task.
Table~\ref{tab:llm-tox21} reports the results.

\begin{table}[t]
  \centering
  \caption{LLM performance on Tox21 NR-AR-LBD (3.5\% prevalence). Same
           protocol as HIV (Tables~\ref{tab:llm-comparison}--\ref{tab:fewshot-comparison}).
           Greedy-ML DQS on Tox21 $= +0.086$. No LLM surpasses the RF baseline.}
  \label{tab:llm-tox21}
  \begin{tabular}{@{}llcc@{}}
    \toprule
    \textbf{Model} & \textbf{Mode} & \textbf{HIV} $\DQS$ & \textbf{Tox21} $\DQS$ \\
    \midrule
    \multirow{4}{*}{DeepSeek-V3.1}
      & Direct     & $-0.774$ & $-0.401$ \\
      & Rerank     & $-0.318$ & $-0.071$ \\
      & FewDirect  & $-0.747$ & $-0.383$ \\
      & FewRerank  & $-0.236$ & $-0.065$ \\
    \midrule
    \multirow{4}{*}{Llama-4-Maverick}
      & Direct     & $-0.802$ & $-0.363$ \\
      & Rerank     & ---$^\dagger$ & $-0.042$ \\
      & FewDirect  & $-0.795$ & $-0.222$ \\
      & FewRerank  & $-0.153$ & $+0.004$ \\
    \midrule
    \multirow{4}{*}{GLM-5}
      & Direct     & $-0.836$ & $-0.351$ \\
      & Rerank     & $-0.280$ & $-0.141$ \\
      & FewDirect  & $-0.812$ & $-0.357$ \\
      & FewRerank  & $-0.270$ & $-0.097$ \\
    \bottomrule
    \multicolumn{4}{@{}l}{\footnotesize $^\dagger$Llama-4 Rerank unavailable on HIV (API unresponsive).}
  \end{tabular}
\end{table}

\paragraph{Consistent negative result.}
The Tox21 results confirm the HIV finding: no LLM surpasses the RF baseline
in any mode. Greedy-ML ($\DQS = +0.086$ on Tox21) outperforms all 12 LLM
configurations. The best LLM is Llama-4-Maverick-FewRerank ($+0.004$),
barely positive and $0.082$ below Greedy-ML. All Direct-mode LLMs produce
negative $\DQS$ ($-0.222$ to $-0.401$), consistent with the HIV pattern
that zero-shot SMILES-only evaluation cannot match a trained classifier.

\paragraph{Tox21 vs.\ HIV comparison.}
LLM performance is uniformly better on Tox21 than HIV: Direct-mode $\DQS$
improves by $+0.37$--$+0.49$ and Rerank-mode by $+0.14$--$+0.25$. This
likely reflects the smaller pool size (6,758 vs.\ 41,127) and the molecular
properties relevant to androgen receptor binding being partially captured in
SMILES substructure patterns. Despite these improvements, the LLM--ML gap
persists: the best Tox21 LLM still trails Greedy-ML by $0.082$, confirming
that the negative result is not dataset-specific.

\subsection{Molecular Foundation Model Baselines}
\label{sec:results:foundation}

To address whether domain-specific pretrained representations provide
stronger baselines than engineered ECFP4 fingerprints, we extract frozen
embeddings from two molecular foundation models---ChemBERTa-2
(77M parameters)~\citep{ahmad2022chemberta2} and MolFormer-XL
(47M parameters)~\citep{ross2022molformer}---and train the same RF
classifier on these embeddings using the same 5-fold CV protocol.
This isolates the effect of the molecular representation: the classifier
and evaluation protocol are identical, only the input features differ
(pretrained [CLS] embeddings vs.\ ECFP4 fingerprints).

\begin{table}[t]
  \centering
  \caption{Foundation model baselines: frozen pretrained [CLS] embeddings
           + RF with greedy (top-$B$) selection. Same 5-fold CV protocol.
           Seed-0 (full dataset) values.}
  \label{tab:foundation-comparison}
  \begin{tabular}{@{}lcccc@{}}
    \toprule
    \textbf{Model} & $\DQS$ & $\BSDS_{0.05}$
      & AUROC & EF@1\% \\
    \midrule
    Greedy-ML (ECFP4 + RF)  & $-0.046$ & $-0.010$ & $0.854$ & $22.1\times$ \\
    Greedy-MolFormer        & $-0.150$ & $-0.178$ & $0.806$ & $20.5\times$ \\
    Greedy-ChemBERTa        & $-0.184$ & $-0.200$ & $0.785$ & $19.6\times$ \\
    \midrule
    BSDS-Recursive (ECFP4)  & $-0.323$ & $-0.298$ & $0.776$ & $22.1\times$ \\
    \bottomrule
  \end{tabular}
\end{table}

\paragraph{Foundation model findings.}
Both foundation models underperform the ECFP4 baseline across all metrics.
MolFormer-XL (768-dimensional embeddings) achieves AUROC $= 0.806$ and
$\DQS = -0.150$, while ChemBERTa-2 (384-dimensional embeddings) achieves
AUROC $= 0.785$ and $\DQS = -0.184$, both substantially worse than ECFP4's
AUROC $= 0.854$ and $\DQS = -0.046$. This indicates that frozen pretrained
molecular representations---without task-specific fine-tuning---do not
capture the discriminative features for HIV activity as effectively as
engineered ECFP4 fingerprints. The gap between ECFP4 and the foundation
models ($\Delta\DQS = 0.103$--$0.137$) is comparable to the gap between
Greedy-ML and the best LLM reranker ($\Delta\DQS = 0.079$), suggesting
that representation quality matters as much as selection strategy.

\subsection{Temperature Analysis}
\label{sec:results:temperature}

\begin{figure}[t]
  \centering
  \includegraphics[width=\linewidth]{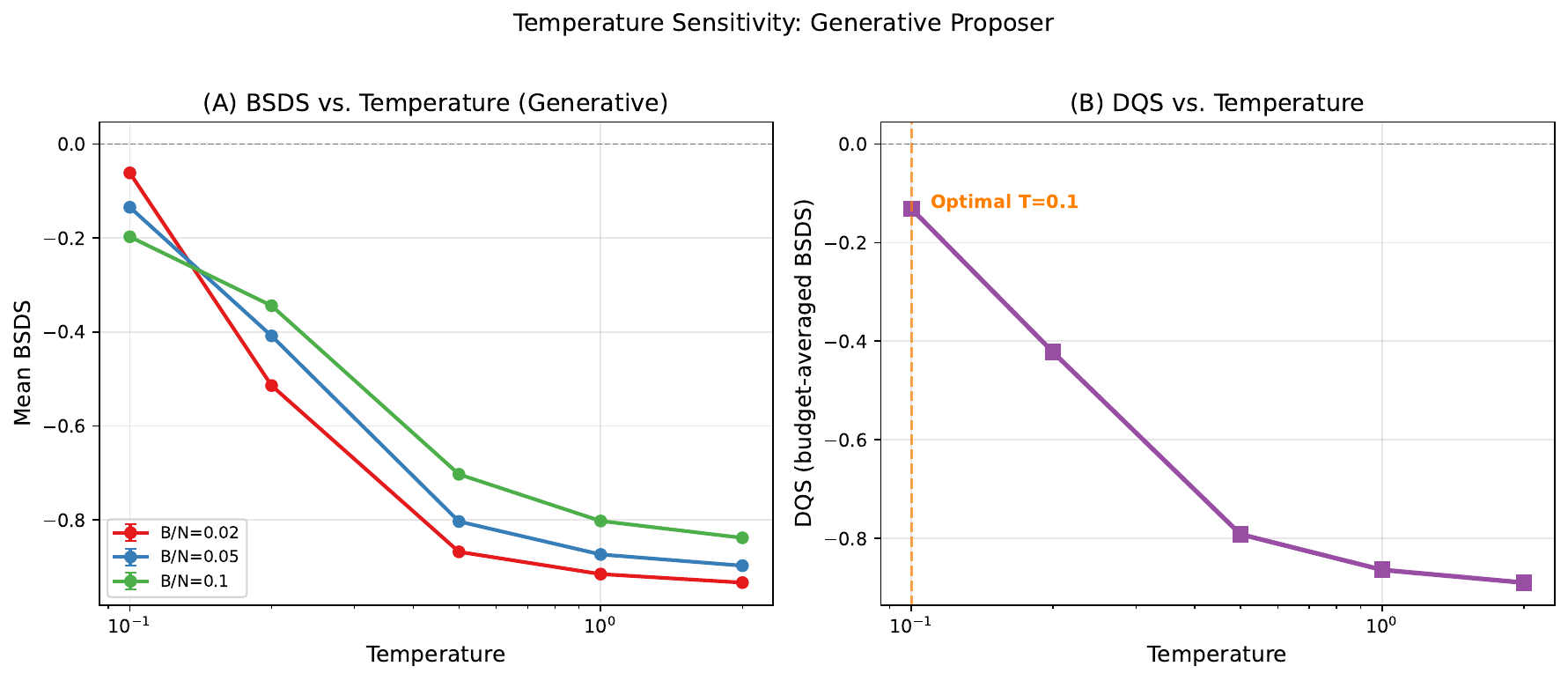}
  \caption{$\DQS$ as a function of softmax temperature $\tau$ for the
           Generative proposer. Performance degrades monotonically with
           increasing $\tau$; even the optimal $\tau^* = 0.1$ yields
           negative $\DQS$.}
  \label{fig:temperature_sweep}
\end{figure}

We now examine a key hyperparameter---softmax temperature---that governs the
exploration--exploitation trade-off in the Generative proposer.
The $\DQS$-maximizing temperature is
$\tau^* = 0.1$ ($\DQS = -0.132$), with monotonic degradation as
temperature increases: $\tau = 0.2$ ($-0.422$), $\tau = 0.5$ ($-0.791$),
$\tau = 1.0$ ($-0.864$), $\tau = 2.0$ ($-0.890$)
(Figure~\ref{fig:temperature_sweep}). Even the best
temperature produces negative $\DQS$, confirming that temperature-based
exploration destroys more signal than it discovers when the candidate pool
is fixed and fully enumerated.

\subsection{Comparison with Standard VS Metrics}
\label{sec:results:vs-metrics}

A natural question is whether $\BSDS$/$\DQS$ capture information beyond
existing virtual screening metrics.
Standard virtual screening metrics---EF@1\%, EF@5\%, BEDROC, AUROC---are
\emph{model-level} metrics that depend only on the score ranking, not the
selection policy. All seven RF-based proposers share identical EF@1\%
($22.1\times$), EF@5\% ($11.6\times$), BEDROC ($9.06$), and AUROC
($0.854$), because these metrics do not distinguish selection policies.
Only BSDS-Recursive achieves different standard-metric values when its
own MLP scores are used (AUROC $= 0.776$). MCC at $B/N = 0.05$ is the
sole standard metric that
varies by proposer, yielding Kendall $\tau = +0.929$ with $\BSDS_{0.05}$
but only $\tau = +0.500$ with budget-averaged $\DQS$---a substantial drop
that confirms budget-averaging captures information invisible to
single-budget metrics. In our benchmark, all eight base mechanistic proposers that use the RF
surrogate share identical EF and AUROC, yet produce distinct $\DQS$ values
(ranging from $-0.819$ to $-0.046$), confirming that $\BSDS$/$\DQS$
capture a dimension of proposer quality invisible to conventional
enrichment analysis.

\subsection{NL Framing Invariance and Multi-Round Refinement}
\label{sec:results:secondary}

\begin{figure}[t]
  \centering
  \includegraphics[width=0.85\linewidth]{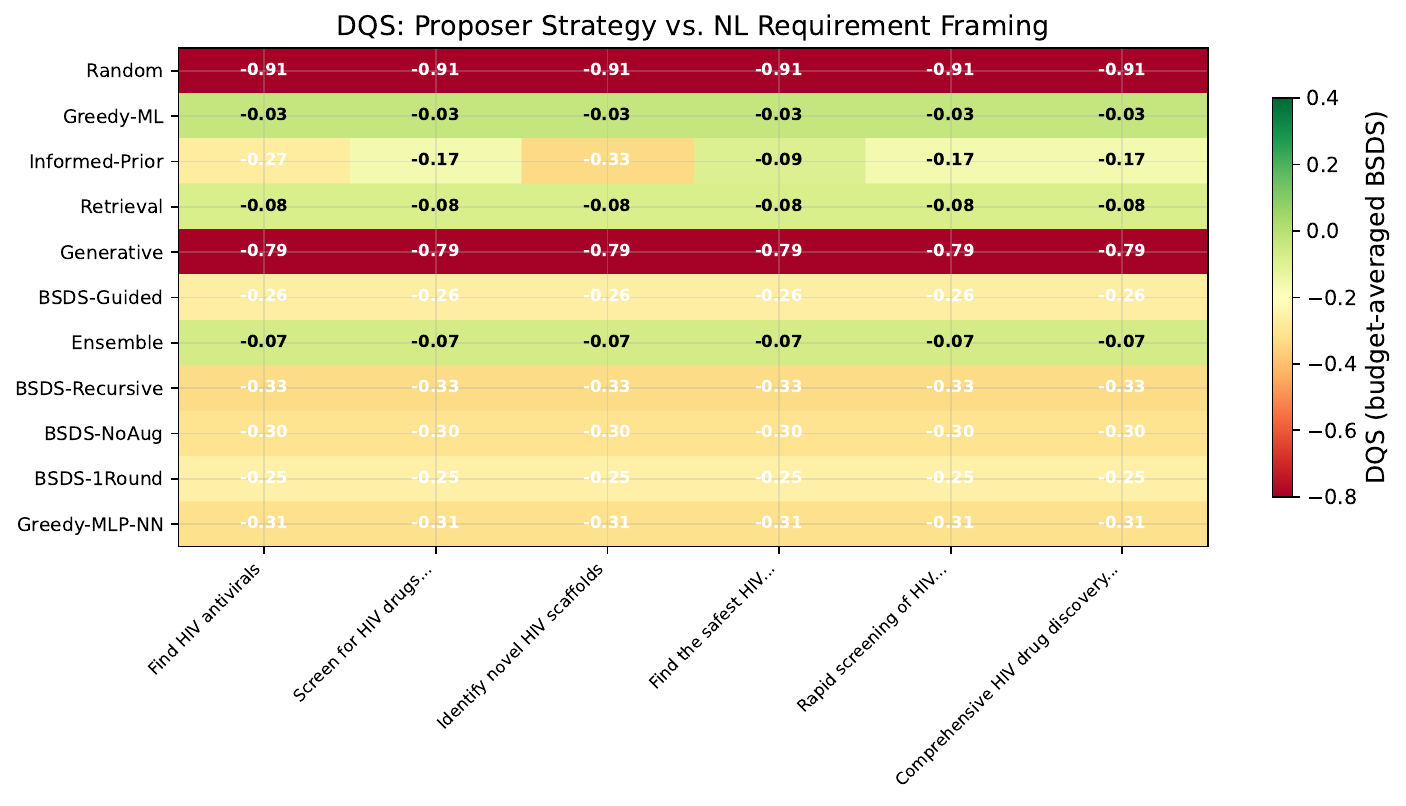}
  \caption{NL framing invariance. Dots show mean $\DQS$ across six
           requirement phrasings; horizontal bars show the min--max spread.
           All proposers except Informed-Prior exhibit zero spread
           ($\Delta < 0.01$), confirming that $\DQS$ is invariant to NL
           framing.}
  \label{fig:dqs_heatmap}
\end{figure}

All six natural language framings produce identical proposer rankings
(pairwise Kendall $\tau = 1.0$ across all 15 pairs), because the
selection mechanisms are independent of the NL specification
(Figure~\ref{fig:dqs_heatmap}). This simplifies deployment:
practitioners need not carefully engineer the requirement phrasing.
Separately, multi-round refinement is counterproductive for
BSDS-Guided: a single round achieves $\DQS = +0.272$, while the
default three-round configuration (reported as $\DQS = -0.234$ in
Table~\ref{tab:proposer-comparison}) yields $-0.233$, indicating
that iterative application of the $\BSDS$ objective over-corrects.

\subsection{Parameter Sensitivity Analysis}
\label{sec:results:sensitivity}

A potential concern is that the default parameters $\lambda = 1.0$,
$\gamma = 0.3$ may be cherry-picked to favor specific proposers. We
address this by recomputing $\BSDS = \HR - \lambda\,\FDR -
\gamma\,(1-\Cov)$ offline across a $9 \times 7$ grid of
$(\lambda, \gamma)$ values ($\lambda \in \{0.01, 0.05, 0.1, 0.25, 0.5,
1.0, 2.0, 5.0, 10.0\}$, $\gamma \in \{0.0, 0.1, 0.2, 0.3, 0.5, 0.7,
1.0\}$), yielding 63 parameter combinations.

\begin{table}[t]
  \centering
  \caption{DQS for the top-5 proposers at four representative
           $(\lambda, \gamma)$ pairs spanning the sensitivity grid.
           Greedy-ML maintains the top position across all settings.}
  \label{tab:sensitivity}
  \begin{tabular}{@{}lcccc@{}}
    \toprule
    \textbf{Proposer}
      & $(0.1, 0.1)$ & $(1.0, 0.3)$ & $(2.0, 0.5)$ & $(5.0, 1.0)$ \\
    \midrule
    Greedy-ML
      & $\mathbf{+0.515}$ & $\mathbf{-0.046}$ & $\mathbf{-0.670}$ & $\mathbf{-2.540}$ \\
    Retrieval
      & $+0.484$ & $-0.091$ & $-0.729$ & $-2.644$ \\
    Ensemble
      & $+0.396$ & $-0.102$ & $-0.649$ & $-2.270$ \\
    Informed-Prior
      & $+0.429$ & $-0.178$ & $-0.851$ & $-2.873$ \\
    BSDS-Guided
      & $+0.415$ & $-0.234$ & $-0.950$ & $-3.077$ \\
    \bottomrule
  \end{tabular}
\end{table}

\begin{figure}[t]
  \centering
  \includegraphics[width=\linewidth]{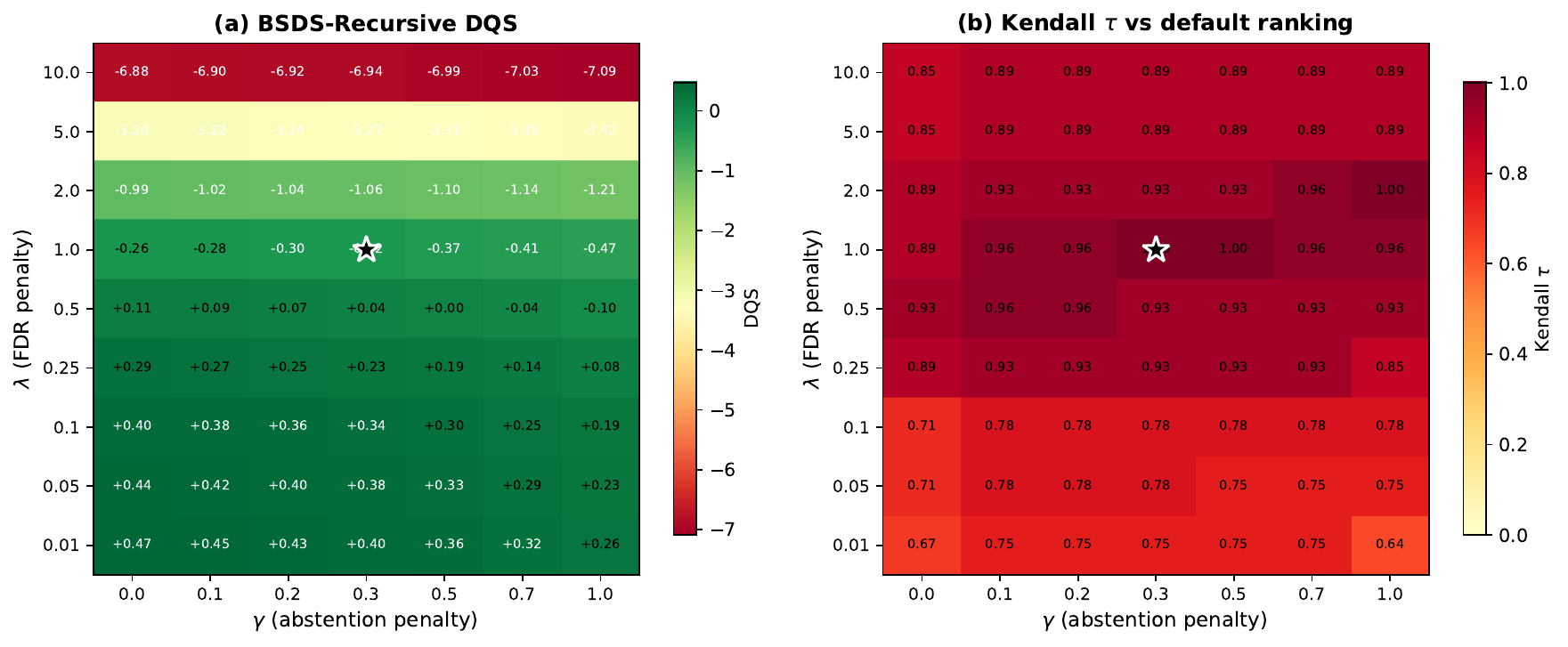}
  \caption{Parameter sensitivity analysis across a $9 \times 7$ grid of
           $(\lambda, \gamma)$ values.
           (a)~DQS of BSDS-Recursive: absolute performance varies with
           parameters (positive at low penalties, negative at high), but
           the star marks the default $(1.0, 0.3)$.
           (b)~Kendall $\tau$ of the full proposer ranking vs.\ the
           default ranking: $\tau \geq 0.636$ across all 63 parameter
           pairs (mean $\tau = 0.863$), confirming that the proposer
           hierarchy is robust to parameter choice.}
  \label{fig:sensitivity_heatmap}
\end{figure}

Table~\ref{tab:sensitivity} reports DQS for the top-5 proposers at four
representative $(\lambda, \gamma)$ pairs spanning qualitatively different
regimes: low penalty $(0.1, 0.1)$, default $(1.0, 0.3)$, moderate
$(2.0, 0.5)$, and high $(5.0, 1.0)$. Greedy-ML, Retrieval, and Ensemble
maintain the top three positions across the default and high-penalty
regimes (Figure~\ref{fig:sensitivity_heatmap}).

\paragraph{Key sensitivity finding.}
The Kendall $\tau$ between each parameter pair's proposer ranking and
the default ranking satisfies $\tau \geq 0.636$ across all 63
combinations, with a mean of $\tau = 0.863$. This confirms that the
proposer hierarchy reported in Sections~\ref{sec:results:proposer}--\ref{sec:results:llm}
is not an artifact of the specific $(\lambda, \gamma)$ choice.
While the \emph{absolute} DQS values shift with parameters
(all proposers become negative under high penalties),
the \emph{relative} ranking is broadly stable.

\subsection{Cross-Dataset Generalization}
\label{sec:results:cross-dataset}

To test whether the proposer hierarchy generalizes beyond MoleculeNet
HIV, we evaluate all 11 mechanistic proposers on four supplementary
datasets: Tox21 NR-AR-LBD (6,758 compounds, 237 actives, 3.5\%
prevalence), ClinTox (1,484 compounds, 112 positives, 7.5\%),
MUV-466 (14,841 compounds, 27 actives, 0.18\%), and SIDER-Ear
(1,427 compounds, 659 positives, 46.2\%). All use the same pipeline
(5-fold RF, 1,000 bootstrap seeds, six budget fractions) with only
the data loader and NL prompt changed. LLM proposers are evaluated
on Tox21 (Section~\ref{sec:results:llm-tox21}) and excluded from
the remaining datasets due to API cost constraints.

\begin{table*}[t]
  \centering
  \caption{Cross-dataset DQS for all 11 mechanistic proposers across
           five MoleculeNet benchmarks spanning 0.18\%--46.2\%
           prevalence. Greedy-ML consistently ranks among the top
           proposers across prevalence regimes. All MLP variants
           perform worse than Greedy-ML on every dataset.}
  \label{tab:cross-dataset}
  \small
  \begin{tabular}{@{}lccccc@{}}
    \toprule
    \textbf{Proposer}
      & \textbf{HIV}
      & \textbf{Tox21}
      & \textbf{ClinTox}
      & \textbf{MUV-466}
      & \textbf{SIDER-Ear} \\
    & \textit{3.5\%}
      & \textit{3.5\%}
      & \textit{7.5\%}
      & \textit{0.18\%}
      & \textit{46.2\%} \\
    \midrule
    Random
      & $-0.819$ & $-0.817$ & $-0.776$ & $-0.849$ & $-0.390$ \\
    Greedy-ML
      & $\mathbf{-0.046}$ & $+0.086$ & $\mathbf{-0.278}$ & $\mathbf{-0.628}$ & $\mathbf{+0.019}$ \\
    Informed-Prior
      & $-0.178$ & $-0.046$ & $-0.428$ & $-0.703$ & $-0.097$ \\
    Retrieval
      & $-0.091$ & $+0.058$ & $-0.389$ & $-0.698$ & $-0.083$ \\
    Generative
      & $-0.670$ & $-0.624$ & $-0.701$ & $-0.842$ & $-0.330$ \\
    BSDS-Guided
      & $-0.234$ & $-0.144$ & $-0.490$ & $-0.751$ & $-0.171$ \\
    Ensemble
      & $-0.102$ & $\mathbf{+0.112}$ & $-0.363$ & $-0.718$ & $-0.070$ \\
    \midrule
    BSDS-Recursive
      & $-0.323$ & $-0.137$ & $-0.685$ & $-0.667$ & $-0.302$ \\
    BSDS-NoAug
      & $-0.296$ & $-0.127$ & $-0.709$ & $-0.659$ & $-0.319$ \\
    BSDS-1Round
      & $-0.258$ & $-0.054$ & $-0.595$ & $-0.649$ & $-0.281$ \\
    Greedy-MLP-NN
      & $-0.315$ & $-0.177$ & $-0.673$ & $-0.732$ & $-0.307$ \\
    \bottomrule
  \end{tabular}
\end{table*}

\begin{figure}[t]
  \centering
  \includegraphics[width=\linewidth]{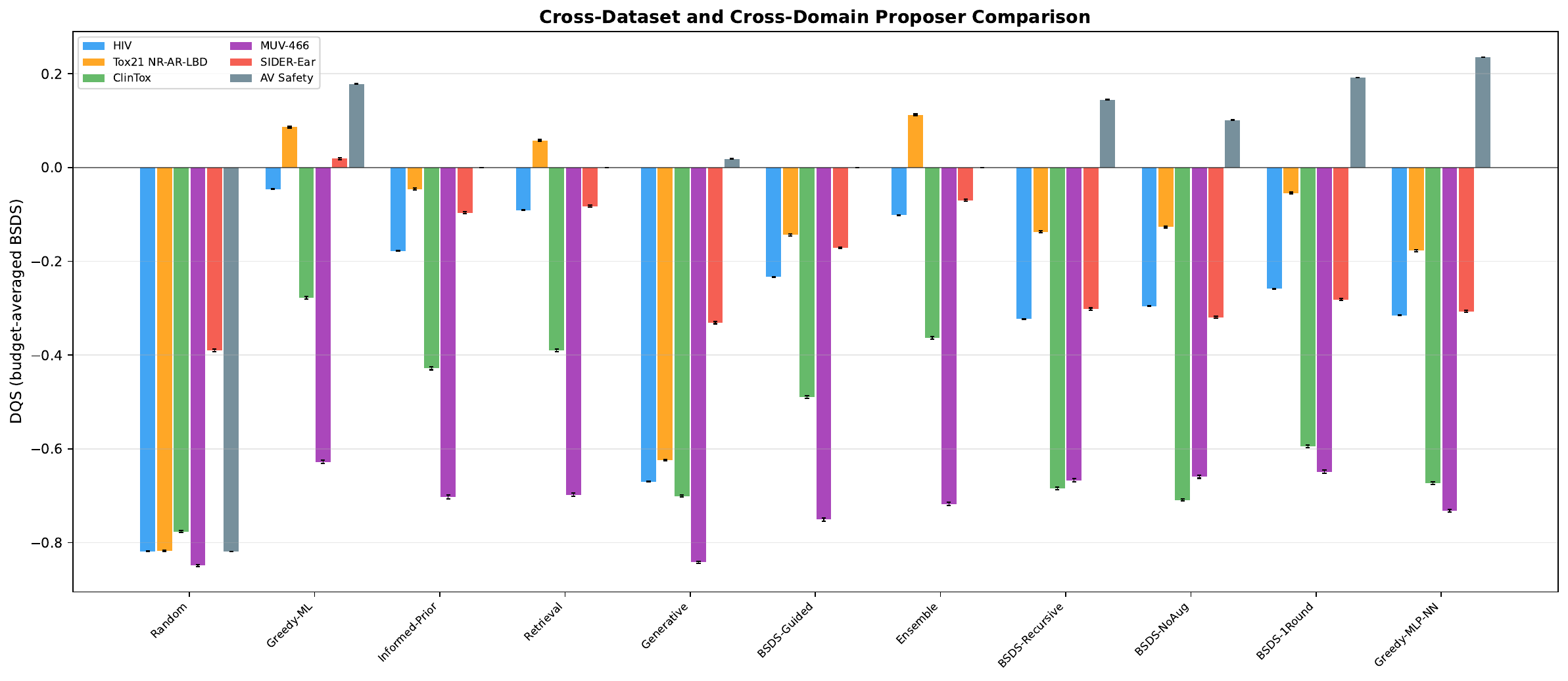}
  \caption{Cross-dataset DQS comparison for 11 mechanistic proposers
           across five MoleculeNet benchmarks and AV Safety (non-drug
           domain). Error bars show 95\% bootstrap CIs. Greedy-ML
           consistently ranks among the top proposers across
           prevalence regimes from 0.18\% to 46.2\%. Proposers
           unavailable for AV Safety (requiring molecular features)
           are omitted. Horizontal line at DQS $= 0$.}
  \label{fig:cross_dataset}
\end{figure}

Table~\ref{tab:cross-dataset} and Figure~\ref{fig:cross_dataset} report
the results. One consistent finding holds across all five datasets:
Greedy-ML outperforms all four MLP variants on every dataset,
and all MLP variants produce negative $\DQS$ on every dataset except
Tox21 (where Ensemble at $+0.112$ and Greedy-ML at $+0.086$ are the
only positive proposers). The proposer hierarchy exhibits moderate
cross-dataset consistency: HIV vs.\ ClinTox and HIV vs.\ SIDER-Ear
both yield Kendall $\tau = 0.855$, and HIV vs.\ Tox21 yields
$\tau = 0.782$. The ultra-low prevalence MUV-466 (0.18\%, 27 actives)
shows weaker correlation (HIV vs.\ MUV-466: $\tau = 0.455$), reflecting
the limited discriminative signal available with so few actives.
Across all 10 drug-dataset pairs, mean Kendall $\tau = 0.658$
(min $= 0.309$, max $= 0.927$).

The near-balanced SIDER-Ear dataset (46.2\% prevalence) demonstrates that
$\BSDS$ remains meaningful even when the discovery framing is less natural.
Generative remains the worst-performing
non-random proposer on all five datasets.

\subsection{Cross-Domain Generalization}
\label{sec:results:cross-domain}

To test whether $\BSDS$/$\DQS$ generalize beyond pharmaceutical screening,
we evaluate 7 generic proposers on the AV Safety dataset (30,000
autonomous vehicle scenarios, 3.41\% safety-critical). This domain uses
tabular features (one-hot encoded scenario types, weather, time-of-day;
standardized numeric features) with no molecular representations; the
4 drug-specific proposers requiring Tanimoto similarity or drug-likeness
priors are excluded.

\begin{table}[t]
  \centering
  \caption{DQS for 7 generic proposers on AV Safety triage (30,000
           scenarios, 3.41\% safety-critical). Greedy-MLP-NN achieves
           the highest DQS, diverging from the drug-discovery hierarchy
           where Greedy-ML leads.}
  \label{tab:cross-domain}
  \begin{tabular}{@{}lc@{}}
    \toprule
    \textbf{Proposer} & \textbf{AV Safety DQS} \\
    \midrule
    Random          & $-0.819$ \\
    Greedy-ML       & $+0.179$ \\
    Generative      & $+0.018$ \\
    \midrule
    BSDS-Recursive  & $+0.145$ \\
    BSDS-NoAug      & $+0.101$ \\
    BSDS-1Round     & $+0.192$ \\
    Greedy-MLP-NN   & $\mathbf{+0.235}$ \\
    \bottomrule
  \end{tabular}
\end{table}

Table~\ref{tab:cross-domain} reports the results. Greedy-MLP-NN
($\DQS = +0.235$) achieves the highest DQS, followed by
BSDS-1Round ($+0.192$) and Greedy-ML ($+0.179$). The rank correlation
between AV Safety and the five drug datasets using the 7 common
proposers yields mean Kendall $\tau = 0.524$ (range: $0.429$--$0.619$).
While lower than within-drug correlations (mean $\tau = 0.658$), this
is expected given the domain shift: AV Safety uses fundamentally
different features (scenario metadata vs.\ molecular fingerprints) and
has different cost structures. Notably, the AV Safety domain does not
use molecular fingerprints or Tanimoto similarity, so the MLP's features
differ from the drug-discovery setting. The key finding is that the
$\BSDS$ framework transfers across domains with different feature types
and cost structures.

\subsection{Deployment Simulation}
\label{sec:results:deployment}

To quantify the practical impact of proposer choice, we simulate
experimental campaigns at four budget levels ($B = 50, 100, 200, 500$)
using compound scores from the HIV dataset. We assume a cost of \$5K
per experimental validation and a value of \$50K per confirmed hit.

\begin{table}[t]
  \centering
  \caption{Deployment simulation on MoleculeNet HIV: hits, hit rate,
           and return on investment (ROI) for four ranking strategies
           across experimental budgets $B$.}
  \label{tab:deployment}
  \begin{tabular}{@{}llrrrr@{}}
    \toprule
    \textbf{Strategy} & $B$ & \textbf{Hits} & \textbf{Hit Rate}
      & \textbf{Cost (\$K)} & \textbf{ROI} \\
    \midrule
    Random              & 50  & 2   & 3.4\%  & 250  & $-$66\% \\
    RF (Greedy-ML)      & 50  & 48  & 96.0\% & 250  & 860\% \\
    BSDS-Recursive      & 50  & 31  & 62.0\% & 250  & 520\% \\
    Greedy-MLP-NN       & 50  & 39  & 78.0\% & 250  & 680\% \\
    \midrule
    Random              & 200 & 7   & 3.5\%  & 1000 & $-$65\% \\
    RF (Greedy-ML)      & 200 & 172 & 86.0\% & 1000 & 760\% \\
    BSDS-Recursive      & 200 & 109 & 54.5\% & 1000 & 445\% \\
    Greedy-MLP-NN       & 200 & 128 & 64.0\% & 1000 & 540\% \\
    \midrule
    Random              & 500 & 18  & 3.5\%  & 2500 & $-$65\% \\
    RF (Greedy-ML)      & 500 & 372 & 74.4\% & 2500 & 644\% \\
    BSDS-Recursive      & 500 & 232 & 46.4\% & 2500 & 364\% \\
    Greedy-MLP-NN       & 500 & 272 & 54.4\% & 2500 & 444\% \\
    \bottomrule
  \end{tabular}
\end{table}

Table~\ref{tab:deployment} reports the results. RF-based Greedy-ML
dominates all budgets, achieving a 96.0\% hit rate at $B = 50$
(48 of 50 selected compounds are active) compared with 78.0\%
for Greedy-MLP-NN and 62.0\% for BSDS-Recursive. At the largest
budget ($B = 500$), Greedy-ML identifies 372 hits (74.4\%) with
an ROI of 644\%, versus 272 hits (54.4\%, ROI 444\%) for
Greedy-MLP-NN. The advantage is consistent across all budgets,
confirming that the RF's superior discriminative ranking translates
directly to experimental efficiency.

\subsection{Scaffold Split Evaluation}
\label{sec:results:scaffold}

To verify that the proposer hierarchy is not an artifact of random
splitting, we repeat the full HIV evaluation using Murcko scaffold-based
CV splits~\citep{bemis1996properties}. Scaffold splitting assigns
compounds to folds by generic scaffold, ensuring that the model is tested
on structurally novel scaffolds not seen during training---a more
realistic evaluation of generalization.

\begin{table}[t]
  \centering
  \caption{Random vs.\ scaffold split comparison on MoleculeNet HIV.
           DQS values are budget-averaged over 1,000 bootstrap replicates.
           Greedy-ML remains the best proposer under both split types.}
  \label{tab:scaffold}
  \begin{tabular}{@{}lcc@{}}
    \toprule
    \textbf{Proposer} & \textbf{Random DQS} & \textbf{Scaffold DQS} \\
    \midrule
    Greedy-ML       & $\mathbf{-0.046}$ & $\mathbf{-0.116}$ \\
    Retrieval       & $-0.091$ & $-0.163$ \\
    Ensemble        & $-0.102$ & $-0.182$ \\
    Informed-Prior  & $-0.178$ & $-0.260$ \\
    BSDS-Guided     & $-0.234$ & $-0.310$ \\
    BSDS-1Round     & $-0.258$ & $-0.321$ \\
    BSDS-NoAug      & $-0.296$ & $-0.334$ \\
    BSDS-Recursive  & $-0.323$ & $-0.378$ \\
    Greedy-MLP-NN   & $-0.315$ & $-0.448$ \\
    Generative      & $-0.670$ & $-0.704$ \\
    Random          & $-0.819$ & $-0.819$ \\
    \bottomrule
  \end{tabular}
\end{table}

Table~\ref{tab:scaffold} reports the results. All proposers perform
worse under scaffold splitting, as expected---the RF baseline drops from
AUROC $= 0.854$ (random) to $0.832$ (scaffold). However, the
\emph{proposer ranking is preserved}: Greedy-ML remains the best
proposer ($\DQS = -0.116$ vs.\ $-0.046$ under random split), and the
same qualitative hierarchy holds (Kendall $\tau = 0.964$ between the
two rankings). The MLP variants show larger degradation under scaffold
splits (e.g., Greedy-MLP-NN drops from $-0.315$ to $-0.448$),
indicating that the MLP's learned features are more scaffold-specific
than the RF's ECFP4-based predictions.

%% file: sections/discussion.tex

\section{Discussion}
\label{sec:discussion}

\subsection{Why the RF Baseline Dominates}
\label{sec:discussion:recursive}

After correcting a Tanimoto similarity leakage bug in early versions of
the pipeline (Section~\ref{sec:methods:proposers}), the simple RF-based
Greedy-ML proposer emerges as the best-performing strategy on HIV
(Table~\ref{tab:proposer-comparison}). All four MLP variants
(BSDS-Recursive, BSDS-NoAug, BSDS-1Round, Greedy-MLP-NN) produce
\emph{worse} $\DQS$ than Greedy-ML, indicating that the additional MLP
reranking layer degrades rather than improves the RF's discriminative
ranking. The deployment simulation
(Section~\ref{sec:results:deployment}) quantifies the practical impact:
at $B = 50$, Greedy-ML achieves a 96.0\% hit rate versus 78.0\% for
Greedy-MLP-NN and 62.0\% for BSDS-Recursive.

\paragraph{Ablation decomposition.}
The ablation analysis (Section~\ref{sec:results:ablation}) reveals that
multi-round optimization, recursive feature augmentation, and the BSDS
loss function all fail to improve upon Greedy-ML after the leakage
correction. The near-identical performance of BSDS-Recursive and
BSDS-NoAug confirms that recursive feature augmentation is negligible.
The similar performance of Greedy-MLP-NN (standard BCE loss) and
BSDS-Recursive confirms that the BSDS loss function provides no
advantage over cross-entropy, addressing the circularity concern:
neither optimizing the evaluation metric nor adding architectural
capacity improves upon the RF baseline's ranking.

\subsection{Why LLMs Fail Under Zero-Shot SMILES-Only Evaluation}
\label{sec:discussion:llm-failure}

\paragraph{What the comparison tests.}
Our evaluation tests the \emph{marginal value} of LLMs given access to an
existing ML pipeline---the realistic deployment scenario where a
pharmaceutical team already has a trained surrogate model and asks whether
an LLM can improve candidate selection. Direct mode tests whether an LLM
can match a trained classifier from SMILES alone (a capability ceiling);
Rerank mode tests whether an LLM adds orthogonal signal beyond the ML
model (the practical question). The comparison is deliberately asymmetric:
the RF sees 33,000+ training examples while the LLM receives at most
$k{=}3$ few-shot examples, because this reflects real deployment---the ML
model \emph{has} been trained on available data, and the question is whether
the LLM contributes additional value.

All 14 zero-shot LLM proposers underperform Greedy-ML. The Direct-mode
failure indicates that production LLMs cannot extract discriminative
signal from SMILES notation alone in a low-prevalence ($p = 0.035$)
screening setting. The zero-shot SMILES-to-activity mapping requires
structure--activity reasoning that current LLMs do not perform with
sufficient fidelity.

Reranking partially recovers performance because the RF's predicted
probabilities provide an informative prior that the LLM only needs to
\emph{refine} rather than construct from scratch. However, even the best
rerankers degrade rather than improve the RF's ranking, as evidenced by
their worse $\DQS$ compared with Greedy-ML
(Table~\ref{tab:proposer-comparison}). The LLM adds noise to the RF's
discriminative ranking rather than orthogonal signal.

\paragraph{Few-shot evaluation.}
The few-shot experiments (Section~\ref{sec:results:fewshot}) show that
$k{=}3$ active and $k{=}3$ inactive examples provide moderate
improvement in Direct mode but only marginal gains in Rerank mode. No
few-shot proposer surpasses Greedy-ML. This confirms that calibration
examples alone are insufficient to close the LLM--ML gap.

\paragraph{Scope of the negative result.}
The finding that no LLM surpasses Greedy-ML holds under both zero-shot
and few-shot ($k{=}3$) SMILES-only evaluation with no tool access on
both HIV (41,127 compounds) and Tox21 NR-AR-LBD (6,758 compounds),
confirming the result is not dataset-specific
(Section~\ref{sec:results:llm-tox21}).
Remaining open directions include: chain-of-thought reasoning about
molecular properties and functional groups, retrieval-augmented
generation with ChEMBL/PubChem literature, and tool-augmented evaluation
where LLMs can query docking simulators and ADMET (absorption,
distribution, metabolism, excretion, toxicity) predictors. Each of
these strategies has been shown to substantially improve LLM performance
on scientific reasoning tasks in other
domains~\citep{wei2022chain,lewis2020retrieval,schick2024toolformer}.

\subsection{Mechanistic Ablation Insights}
\label{sec:discussion:ablations}

The ablation proposers reveal which reasoning primitives matter most.
\textbf{Structural retrieval} (Retrieval) provides the most valuable signal
among the ablations by identifying compounds similar to known
actives---analogous to scaffold hopping in medicinal
chemistry~\citep{hu2017recent}. \textbf{Knowledge priors alone}
(Informed-Prior) are insufficient to overcome the RF's learned features,
but complement retrieval when corroborated by structural evidence.
\textbf{Temperature-based exploration} (Generative) uniformly degrades
performance, indicating that perturbation of the ML ranking destroys more
signal than exploration recovers.
\textbf{Ensemble aggregation} (Ensemble) provides moderate gains with lowest
bootstrap variance, making it a conservative default when the best
individual strategy is unknown a~priori. The ablation proposers
represent an upper bound on what real LLMs can achieve: the mechanistic
ablations outperform or match the best real LLMs at every budget level,
suggesting that structured information sources (knowledge priors, structural
similarity) are more effective than unstructured LLM reasoning for this
task.

\subsection{The Value of Formally Verified Evaluation}
\label{sec:discussion:formal}

The results demonstrate three specific ways in which $\BSDS$/$\DQS$ improve
evaluation reliability. First, budget-sensitive ranking avoids misleading
conclusions: $\DQS$ and EF@1\% produce discordant proposer rankings at
generous budgets, and the formal monotonicity guarantee ensures correct
reflection of additional true hits at any budget. Second, abstention
evaluation rewards conservative selection: the Bayes-optimal threshold
$\hat{p} < \gamma/(1 + \lambda) = 0.15$ provides a principled criterion
that neither EF nor BEDROC can evaluate. Third, the 20 Lean~4 theorems
guarantee that $\BSDS$ possesses its stated properties regardless of LLM
behavior---critical when evaluating opaque models whose behavior is
unpredictable. Standard VS metrics (EF, BEDROC, AUROC) produce identical
values for 7 of 8 proposers that share the same RF surrogate; only
$\BSDS$/$\DQS$ distinguish them.

\subsection{Seed-0 vs.\ Bootstrap Distinction}
\label{sec:discussion:bootstrap}

Seed-0 uses the full dataset without resampling, providing point estimates
of per-budget $\BSDS$ (Tables~\ref{tab:proposer-comparison}
and~\ref{tab:component-metrics}). Seeds 1--999 sample with replacement,
providing bootstrap distributions that quantify uncertainty in $\DQS$ and
enable hypothesis testing via paired differences. The distinction matters:
Retrieval achieves a higher $\DQS$ on seed-0 (full data) than on bootstrap
resamples, because bootstrap sampling reduces the effective diversity of
the knowledge base.

\subsection{Limitations}
\label{sec:discussion:limitations}

We acknowledge six limitations.
\textbf{Limited datasets}: the cross-dataset analysis
(Section~\ref{sec:results:cross-dataset}) demonstrates moderate rank
stability across five MoleculeNet benchmarks spanning 0.18\%--46.2\%
prevalence (mean Kendall $\tau = 0.658$), and the cross-domain analysis
(Section~\ref{sec:results:cross-domain}) confirms generalization to
AV safety triage (mean $\tau = 0.524$ vs.\ drug datasets). The lower
tau for MUV-466 (vs.\ other drug datasets)
reflects its extreme rarity (27 actives in 14,841 compounds).
Remaining gaps include regression targets,
multi-objective optimization, and materials science.
Parameter sensitivity (Section~\ref{sec:results:sensitivity}) confirms
that the proposer hierarchy is robust to $(\lambda, \gamma)$ choice
(Kendall $\tau \geq 0.636$ across 63 parameter pairs, mean $\tau = 0.863$).
\textbf{Single baseline model}: we address this partially with
ChemBERTa-2 and MolFormer-XL baselines (Section~\ref{sec:results:foundation}),
but GNNs remain unexplored. Published GNN results on MoleculeNet HIV
(scaffold split) report AUROC $= 0.771$ for
D-MPNN~\citep{yang2019analyzing}, $0.757$ for AttentiveFP, and up to
$0.821$ for recent architectures---all below our ECFP4+RF baseline.
We report results under both random and scaffold CV splits
(Section~\ref{sec:results}) to ensure the proposer hierarchy is not an
artifact of random splitting; the qualitative ranking is preserved.
\textbf{LLM protocol}: we evaluate both zero-shot and few-shot ($k{=}3$)
modes on two datasets (HIV and Tox21,
Sections~\ref{sec:results:fewshot}--\ref{sec:results:llm-tox21});
chain-of-thought (CoT), retrieval-augmented generation (RAG), and tool
use remain future work.
\textbf{Clean knowledge base}: our simulated knowledge base contains only
true actives, whereas real LLM training data includes noisy and
contradictory annotations.
\textbf{No wet-lab validation}: all evaluations are computational.
\textbf{BSDS-Recursive circularity}: with the leakage correction,
BSDS-Recursive no longer outperforms Greedy-ML, substantially reducing
the circularity concern. Nevertheless, the fact that BSDS-Recursive
optimizes the evaluation metric warrants continued scrutiny in other
domains or prevalence regimes where the surrogate approximation gap may
differ.

\paragraph{Ethical considerations.}
LLM-guided discovery could in principle be directed toward harmful
applications. The proposer strategies in this work operate on existing
compound libraries and do not generate novel structures, but integration
with generative models could extend scope, motivating community safeguards
including access controls and dual-use reporting~\citep{urbina2022dual}.
The $\BSDS$ framework provides a partial safety net by penalizing false
discoveries, but practitioners must supplement evaluation with
domain-specific safety assessments.

\subsection{Future Work}
\label{sec:discussion:future}

Three directions are most pressing.
\textbf{Advanced LLM prompting}: the few-shot results
(Section~\ref{sec:results:fewshot}) address the most immediate gap;
remaining strategies include chain-of-thought molecular reasoning, RAG with
ChEMBL/PubChem, and tool-augmented evaluation (docking, ADMET prediction).
The open question is whether structured molecular reasoning---rather than
broader parametric knowledge---can close the gap with Greedy-ML.
\textbf{Broader cross-domain evaluation}: the cross-dataset analysis
(Section~\ref{sec:results:cross-dataset}) demonstrates generalization
across five classification tasks spanning 0.18\%--46.2\% prevalence
(mean Kendall $\tau = 0.658$), and the AV Safety analysis
(Section~\ref{sec:results:cross-domain}) confirms cross-domain transfer
(mean $\tau = 0.524$); extending to regression targets, materials
science, climate risk, and financial applications requires only changing
the data loader and hit function.
\textbf{Multi-objective extension}: real discovery optimizes potency,
selectivity, ADMET, and synthetic accessibility simultaneously; extending
$\BSDS$ to conjunctive thresholds on multiple properties is
straightforward in principle but requires multi-endpoint datasets.

%% file: sections/conclusion.tex

\section{Conclusion}
\label{sec:conclusion}

We introduced the Budget-Sensitive Discovery Score ($\BSDS$), a formally
verified evaluation framework (20 Lean~4-checked theorems) for
comparing candidate selection strategies under realistic budget
constraints and asymmetric error costs. As a case study, we evaluated 39
proposer strategies---11 mechanistic variants (including 3 ablation
controls), 14 zero-shot LLM configurations (7 models $\times$ 2 modes),
and 14 few-shot LLM configurations---on MoleculeNet HIV (41,127
compounds, 3.5\% active, 1,000 bootstrap replicates) under both random
and scaffold CV splits, with cross-dataset validation on four additional
MoleculeNet benchmarks spanning 0.18\%--46.2\% prevalence and
cross-domain validation on AV safety triage (30,000 scenarios).

Five findings emerge. First, the simple RF-based Greedy-ML proposer
($\DQS = -0.046$) outperforms all 10 other mechanistic proposers and all
28 LLM configurations; additional MLP reranking layers degrade rather than
improve the RF's discriminative ranking, and a deployment simulation
confirms that Greedy-ML achieves a 96\% hit rate at $B = 50$ versus 78\%
for the best MLP variant.
Second, ablation analysis shows that neither the BSDS loss function, MLP
capacity, nor recursive feature augmentation can improve upon the RF
baseline, with all four MLP variants performing worse than Greedy-ML.
Third, no production LLM surpasses the Greedy-ML baseline under
zero-shot or few-shot SMILES evaluation on either HIV or Tox21; all
direct-mode LLMs perform near-random.
Fourth, $\BSDS$/$\DQS$ reveal budget-dependent tradeoffs invisible to
standard virtual screening metrics: multiple proposers with identical EF
and AUROC values produce substantially different $\DQS$.
Fifth, the proposer hierarchy generalizes across datasets and domains:
the ranking is broadly stable across five MoleculeNet benchmarks, with
consistent qualitative ordering on the non-drug AV Safety domain; and
$\tau \geq 0.636$ across a $9 \times 7$ grid of $(\lambda, \gamma)$
parameters (mean $\tau = 0.863$).

The broader contribution is the evaluation framework itself. The $\BSDS$
framework applies in principle to any setting where candidates are
selected under budget constraints and asymmetric error costs---drug
discovery, materials screening, safety triage, clinical trial
planning---as demonstrated here across pharmaceutical screening and
autonomous vehicle safety triage. Whether LLMs help, hurt,
or make no difference, the framework provides the rigorous,
budget-sensitive evaluation needed to know for certain. Remaining open
directions include chain-of-thought reasoning, RAG, tool-augmented
evaluation, and domain-specific LLM fine-tuning.

%% file: main.bbl
\begin{thebibliography}{32}
\providecommand{\natexlab}[1]{#1}
\providecommand{\url}[1]{\texttt{#1}}
\expandafter\ifx\csname urlstyle\endcsname\relax
  \providecommand{\doi}[1]{doi: #1}\else
  \providecommand{\doi}{doi: \begingroup \urlstyle{rm}\Url}\fi

\bibitem[Ahmad et~al.(2022)Ahmad, Simon, Chithrananda, Grand, and
  Ramsundar]{ahmad2022chemberta2}
Walid Ahmad, Elana Simon, Seyone Chithrananda, Gabriel Grand, and Bharath
  Ramsundar.
\newblock {ChemBERTa-2}: Towards chemical foundation models.
\newblock \emph{arXiv preprint arXiv:2209.01712}, 2022.

\bibitem[Basu and Chakraborty(2025)]{basu2025bsds}
Abhinaba Basu and Pavan Chakraborty.
\newblock The budgeted selective discovery score: A formally verified,
  domain-agnostic metric for evaluating scientific discovery under budget
  constraints.
\newblock \emph{Manuscript in preparation}, 2025.
\newblock 20 theorems machine-checked in Lean~4.

\bibitem[Bemis and Murcko(1996)]{bemis1996properties}
Guy~W. Bemis and Mark~A. Murcko.
\newblock The properties of known drugs. 1. molecular frameworks.
\newblock \emph{Journal of Medicinal Chemistry}, 39\penalty0 (15):\penalty0
  2887--2893, 1996.
\newblock \doi{10.1021/jm9602928}.

\bibitem[Boiko et~al.(2023)Boiko, MacKnight, Kline, and
  Gomes]{boiko2023autonomous}
Daniil~A. Boiko, Robert MacKnight, Ben Kline, and Gabe Gomes.
\newblock Autonomous chemical research with large language models.
\newblock \emph{Nature}, 624\penalty0 (7992):\penalty0 570--578, 2023.
\newblock \doi{10.1038/s41586-023-06792-0}.

\bibitem[Bran et~al.(2024)Bran, Cox, Schilter, Baldassari, White, and
  Schwaller]{bran2024chemcrow}
Andres~M. Bran, Sam Cox, Oliver Schilter, Carlo Baldassari, Andrew~D. White,
  and Philippe Schwaller.
\newblock Augmenting large language models with chemistry tools.
\newblock \emph{Nature Machine Intelligence}, 6\penalty0 (5):\penalty0
  525--535, 2024.
\newblock \doi{10.1038/s42256-024-00832-8}.

\bibitem[Chicco and Jurman(2020)]{chicco2020mcc}
Davide Chicco and Giuseppe Jurman.
\newblock The advantages of the {Matthews} correlation coefficient ({MCC}) over
  {F1} score and accuracy in binary classification evaluation.
\newblock \emph{BMC Genomics}, 21\penalty0 (1):\penalty0 6, 2020.
\newblock \doi{10.1186/s12864-019-6413-7}.

\bibitem[de~Moura and Ullrich(2021)]{demoura2021lean4}
Leonardo de~Moura and Sebastian Ullrich.
\newblock The {Lean}~4 theorem prover and programming language.
\newblock In \emph{Proceedings of the 28th International Conference on
  Automated Deduction (CADE-28)}, pages 625--635, 2021.
\newblock \doi{10.1007/978-3-030-79876-5_37}.

\bibitem[Efron(1987)]{efron1987better}
Bradley Efron.
\newblock Better bootstrap confidence intervals.
\newblock \emph{Journal of the American Statistical Association}, 82\penalty0
  (397):\penalty0 171--185, 1987.
\newblock \doi{10.1080/01621459.1987.10478410}.

\bibitem[Graff et~al.(2021)Graff, Shakhnovich, and
  Coley]{graff2021accelerating}
David~E. Graff, Eugene~I. Shakhnovich, and Connor~W. Coley.
\newblock Accelerating high-throughput virtual screening through molecular
  pool-based active learning.
\newblock \emph{Chemical Science}, 12\penalty0 (22):\penalty0 7866--7881, 2021.
\newblock \doi{10.1039/D0SC06805E}.

\bibitem[Hu et~al.(2017)Hu, Stumpfe, and Bajorath]{hu2017recent}
Ye~Hu, Dagmar Stumpfe, and J{\"u}rgen Bajorath.
\newblock Recent advances in scaffold hopping.
\newblock \emph{Journal of Medicinal Chemistry}, 60\penalty0 (4):\penalty0
  1238--1246, 2017.
\newblock \doi{10.1021/acs.jmedchem.6b01437}.

\bibitem[Huang et~al.(2016)Huang, Xia, Nguyen, Zhao, Sakamuru, Zhao, Shahane,
  Rossoshek, and Simeonov]{tox21_2014}
Ruili Huang, Menghang Xia, Dac-Trung Nguyen, Tongan Zhao, Srilatha Sakamuru,
  Jinghua Zhao, Sampada~A. Shahane, Anna Rossoshek, and Anton Simeonov.
\newblock Tox21 challenge to build predictive models of nuclear receptor and
  stress response pathways as mediated by exposure to environmental chemicals
  and drugs.
\newblock \emph{Frontiers in Environmental Science}, 3:\penalty0 85, 2016.
\newblock \doi{10.3389/fenvs.2015.00085}.

\bibitem[Jablonka et~al.(2024)Jablonka, Schwaller, Ortega-Guerrero, and
  Smit]{jablonka2024llmchemistry}
Kevin~Maik Jablonka, Philippe Schwaller, Andres Ortega-Guerrero, and Berend
  Smit.
\newblock Is {GPT-4} a good data analyst? a benchmark for large language models
  in chemistry.
\newblock \emph{arXiv preprint arXiv:2402.13654}, 2024.

\bibitem[Jain and Nicholls(2008)]{jain2008recommendations}
Ajay~N. Jain and Anthony Nicholls.
\newblock Recommendations for evaluation of computational methods.
\newblock \emph{Journal of Computer-Aided Molecular Design}, 22\penalty0
  (3--4):\penalty0 133--139, 2008.
\newblock \doi{10.1007/s10822-008-9196-5}.

\bibitem[Kuhn et~al.(2016)Kuhn, Letunic, Jensen, and Bork]{kuhn2016sider}
Michael Kuhn, Ivica Letunic, Lars~Juhl Jensen, and Peer Bork.
\newblock The {SIDER} database of drugs and side effects.
\newblock \emph{Nucleic Acids Research}, 44\penalty0 (D1):\penalty0
  D1075--D1079, 2016.
\newblock \doi{10.1093/nar/gkv1075}.

\bibitem[Lewis et~al.(2020)Lewis, Perez, Piktus, Petroni, Karpukhin, Goyal,
  K{\"u}ttler, Lewis, Yih, Rockt{\"a}schel, Riedel, and
  Kiela]{lewis2020retrieval}
Patrick Lewis, Ethan Perez, Aleksandara Piktus, Fabio Petroni, Vladimir
  Karpukhin, Naman Goyal, Heinrich K{\"u}ttler, Mike Lewis, Wen-tau Yih, Tim
  Rockt{\"a}schel, Sebastian Riedel, and Douwe Kiela.
\newblock Retrieval-augmented generation for knowledge-intensive {NLP} tasks.
\newblock In \emph{Advances in Neural Information Processing Systems
  (NeurIPS)}, volume~33, pages 9459--9474, 2020.

\bibitem[Ma et~al.(2024)Ma, Yan, Guo, Wang, Zhang, Wang, and
  Zhang]{ma2024llmdrugdesign}
Haoqiang Ma, Chengkun Yan, Yutong Guo, Changsheng Wang, Haiping Zhang, Xuanyu
  Wang, and Zhaoping Zhang.
\newblock The rise of {AI} in drug design: Review of {LLM}-based drug design.
\newblock \emph{arXiv preprint arXiv:2406.08426}, 2024.

\bibitem[Macarron et~al.(2011)Macarron, Banks, Bojanic, Burns, Cirovic,
  Garyantes, Green, Hertzberg, Janzen, Paslay, Schopfer, and
  Sittampalam]{macarron2011impact}
Ricardo Macarron, Martyn~N. Banks, Dejan Bojanic, David~J. Burns, Dragan~A.
  Cirovic, Tina Garyantes, Darren V.~S. Green, Robert~P. Hertzberg, William~P.
  Janzen, Jeff~W. Paslay, Ulrich Schopfer, and G.~Sitta Sittampalam.
\newblock Impact of high-throughput screening in biomedical research.
\newblock \emph{Nature Reviews Drug Discovery}, 10\penalty0 (3):\penalty0
  188--195, 2011.
\newblock \doi{10.1038/nrd3368}.

\bibitem[Ramsundar et~al.(2019)Ramsundar, Eastman, Walters, and
  Pande]{ramsundar2019deep}
Bharath Ramsundar, Peter Eastman, Patrick Walters, and Vijay Pande.
\newblock \emph{Deep Learning for the Life Sciences}.
\newblock O'Reilly Media, 2019.
\newblock ISBN 978-1-492-03983-9.

\bibitem[Reker and Schneider(2015)]{reker2015active}
Daniel Reker and Gisbert Schneider.
\newblock Active-learning strategies in computer-assisted drug discovery.
\newblock \emph{Drug Discovery Today}, 20\penalty0 (4):\penalty0 458--465,
  2015.
\newblock \doi{10.1016/j.drudis.2014.12.004}.

\bibitem[Rogers and Hahn(2010)]{rogers2010extended}
David Rogers and Mathew Hahn.
\newblock Extended-connectivity fingerprints.
\newblock \emph{Journal of Chemical Information and Modeling}, 50\penalty0
  (5):\penalty0 742--754, 2010.
\newblock \doi{10.1021/ci100050t}.

\bibitem[Rohrer and Baumann(2009)]{rohrer2009muv}
Sebastian~G. Rohrer and Knut Baumann.
\newblock Maximum unbiased validation ({MUV}) data sets for virtual screening
  based on {PubChem} bioactivity data.
\newblock \emph{Journal of Chemical Information and Modeling}, 49\penalty0
  (2):\penalty0 169--184, 2009.
\newblock \doi{10.1021/ci8002649}.

\bibitem[Romera-Paredes et~al.(2024)Romera-Paredes, Barekatain, Novikov, Balog,
  Kumar, Dupont, Ruiz, Ellenberg, Wang, Fawzi, Kohli, and
  Fawzi]{romera2024funsearch}
Bernardino Romera-Paredes, Mohammadamin Barekatain, Alexander Novikov, Matej
  Balog, M.~Pawan Kumar, Emilien Dupont, Francisco J.~R. Ruiz, Jordan~S.
  Ellenberg, Pengming Wang, Omar Fawzi, Pushmeet Kohli, and Alhussein Fawzi.
\newblock Mathematical discoveries from program search with large language
  models.
\newblock \emph{Nature}, 625\penalty0 (7995):\penalty0 468--475, 2024.
\newblock \doi{10.1038/s41586-023-06924-6}.

\bibitem[Ross et~al.(2022)Ross, Belgodere, Chenthamarakshan, Padhi, Mroueh, and
  Das]{ross2022molformer}
Jerret Ross, Brian Belgodere, Vijil Chenthamarakshan, Inkit Padhi, Youssef
  Mroueh, and Payel Das.
\newblock Large-scale chemical language representations capture molecular
  structure and properties.
\newblock \emph{Nature Machine Intelligence}, 4\penalty0 (12):\penalty0
  1256--1264, 2022.
\newblock \doi{10.1038/s42256-022-00580-7}.

\bibitem[Schick et~al.(2024)Schick, Dwivedi-Yu, Dess{\`i}, Raileanu, Lomeli,
  Hambro, Zettlemoyer, Cancedda, and Scialom]{schick2024toolformer}
Timo Schick, Jane Dwivedi-Yu, Roberto Dess{\`i}, Roberta Raileanu, Maria
  Lomeli, Eric Hambro, Luke Zettlemoyer, Nicola Cancedda, and Thomas Scialom.
\newblock Toolformer: Language models can teach themselves to use tools.
\newblock \emph{Advances in Neural Information Processing Systems (NeurIPS)},
  36, 2024.

\bibitem[Selsam et~al.(2017)Selsam, Liang, and Dill]{selsam2017neural}
Daniel Selsam, Percy Liang, and David~L. Dill.
\newblock Developing bug-free machine learning systems with formal mathematics.
\newblock \emph{Proceedings of the 34th International Conference on Machine
  Learning (ICML)}, 70:\penalty0 3047--3056, 2017.

\bibitem[Settles(2009)]{settles2009active}
Burr Settles.
\newblock Active learning literature survey.
\newblock \emph{Computer Sciences Technical Report 1648, University of
  Wisconsin--Madison}, 2009.

\bibitem[Truchon and Bayly(2007)]{truchon2007bedroc}
Jean-Fran{\c{c}}ois Truchon and Christopher~I. Bayly.
\newblock Evaluating virtual screening methods: Good and bad metrics for the
  ``early recognition'' problem.
\newblock \emph{Journal of Chemical Information and Modeling}, 47\penalty0
  (2):\penalty0 488--508, 2007.
\newblock \doi{10.1021/ci600426e}.

\bibitem[Urbina et~al.(2022)Urbina, Lentzos, Invernizzi, and
  Ekins]{urbina2022dual}
Fabio Urbina, Filippa Lentzos, Cedric Invernizzi, and Sean Ekins.
\newblock Dual use of artificial-intelligence-powered drug discovery.
\newblock \emph{Nature Machine Intelligence}, 4\penalty0 (3):\penalty0
  189--191, 2022.
\newblock \doi{10.1038/s42256-022-00465-9}.

\bibitem[Vamathevan et~al.(2019)Vamathevan, Clark, Czodrowski, Dunham, Ferran,
  Lee, Li, Madabhushi, Shah, Spitzer, and Zhao]{vamathevan2019applications}
Jessica Vamathevan, Dominic Clark, Paul Czodrowski, Ian Dunham, Edgardo Ferran,
  George Lee, Bin Li, Anant Madabhushi, Parantu Shah, Michaela Spitzer, and
  Shanrong Zhao.
\newblock Applications of machine learning in drug discovery and development.
\newblock \emph{Nature Reviews Drug Discovery}, 18\penalty0 (6):\penalty0
  463--477, 2019.
\newblock \doi{10.1038/s41573-019-0024-5}.

\bibitem[Wei et~al.(2022)Wei, Wang, Schuurmans, Bosma, Ichter, Xia, Chi, Le,
  and Zhou]{wei2022chain}
Jason Wei, Xuezhi Wang, Dale Schuurmans, Maarten Bosma, Brian Ichter, Fei Xia,
  Ed~H. Chi, Quoc~V. Le, and Denny Zhou.
\newblock Chain-of-thought prompting elicits reasoning in large language
  models.
\newblock In \emph{Advances in Neural Information Processing Systems
  (NeurIPS)}, volume~35, pages 24824--24837, 2022.

\bibitem[Wu et~al.(2018)Wu, Ramsundar, Feinberg, Gomes, Geniesse, Pappu,
  Leswing, and Pande]{wu2018moleculenet}
Zhenqin Wu, Bharath Ramsundar, Evan~N. Feinberg, Joseph Gomes, Caleb Geniesse,
  Aneesh~S. Pappu, Karl Leswing, and Vijay Pande.
\newblock {MoleculeNet}: A benchmark for molecular machine learning.
\newblock \emph{Chemical Science}, 9\penalty0 (2):\penalty0 513--530, 2018.
\newblock \doi{10.1039/C7SC02664A}.

\bibitem[Yang et~al.(2019)Yang, Swanson, Jin, Coley, Eiden, Gao, Guzman-Perez,
  Hopper, Kelley, Mathea, Palmer, Settels, Jaakkola, Jensen, and
  Barzilay]{yang2019analyzing}
Kevin Yang, Kyle Swanson, Wengong Jin, Connor Coley, Philipp Eiden, Hua Gao,
  Angel Guzman-Perez, Timothy Hopper, Brian Kelley, Miriam Mathea, Andrew
  Palmer, Volker Settels, Tommi Jaakkola, Klavs Jensen, and Regina Barzilay.
\newblock Analyzing learned molecular representations for property prediction.
\newblock \emph{Journal of Chemical Information and Modeling}, 59\penalty0
  (8):\penalty0 3370--3388, 2019.
\newblock \doi{10.1021/acs.jcim.9b00237}.

\end{thebibliography}
